\RequirePackage{fix-cm}
\documentclass[twocolumn]{svjour3}          
\smartqed  
\usepackage{graphicx}
\graphicspath{{figures/}}
\DeclareGraphicsExtensions{.pdf,.jpeg,.png}
\usepackage{mathtools,commath,amssymb,mathrsfs,stmaryrd}
\usepackage{tabularx}
\usepackage{multirow}
\usepackage{tcolorbox}\tcbset{before skip=1ex, after skip=1ex, left=1ex, right=1ex}
\usepackage{url}
\usepackage[linesnumbered]{algorithm2e}

\def\vec#1{\mathchoice{\mbox{\boldmath$\displaystyle#1$}}
  {\mbox{\boldmath$\textstyle#1$}}
  {\mbox{\boldmath$\scriptstyle#1$}}
  {\mbox{\boldmath$\scriptscriptstyle#1$}}}
\newcommand{\mat}[1]{\mathbf{#1}}
\newcommand{\argminD}[1]{\underset{#1}{\arg\!\min}} 

\newcommand{\definedas}{\overset{\mathrm{def}}{=}}
\newcommand{\nSV}{n_\text{sv}}

\newtheorem{Definition}{Definition}
\begin{document}

\title{Human Pose and Path Estimation from Aerial Video using Dynamic Classifier Selection
}

\author{Asanka G Perera \and Yee Wei Law \and Javaan Chahl}



\institute{A. G. Perera \and Y. W. Law \and J. Chahl 
              \at School of Engineering, University of South Australia, Mawson Lakes, SA 5095, Australia \\
              \email{asanka.perera@mymail.unisa.edu.au}           
           \and
           Y. W. Law  \at
              \email{yeewei.law@unisa.edu.au}
           \and
           J. Chahl
               \at Joint and Operations Analysis Division, Defence Science and Technology Group, Melbourne, Victoria 3207, Australia \\
              \email{javaan.chahl@unisa.edu.au}
}

\date{Received: date / Accepted: date}

\maketitle

\begin{abstract}
\textbf{Background / introduction}--- We consider the problem of estimating human pose and trajectory by an aerial robot with a monocular camera in near real time. We present a preliminary solution whose distinguishing feature is a \emph{dynamic classifier selection} architecture. 

\textbf{Methods}--- In our solution, each video frame is corrected for perspective using projective transformation. Then, two alternative feature sets are used: (i) Histogram of Oriented Gradients (HOG) of the silhouette, (ii) Convolutional Neural Network (CNN) features of the RGB image. The features (HOG or CNN) are classified using a \emph{dynamic classifier}. A class is defined as a pose-viewpoint pair, and a total of 64 classes are defined to represent a forward walking and turning gait sequence. 

\textbf{Results}--- Our solution provides three main advantages: (i) Classification is efficient due to dynamic selection (4-class vs. 64-class classification). (ii) Classification errors are confined to neighbors of the true viewpoints. (iii) The robust temporal relationship between poses is used to resolve the left-right ambiguities of human silhouettes. 

\textbf{Conclusions}--- Experiments conducted on both fronto-parallel videos and aerial videos confirm our solution can achieve accurate pose and trajectory estimation for both scenarios. We found using HOG features provides higher accuracy than using CNN features. For example, applying the HOG-based variant of our scheme to the ``walking on a figure 8-shaped path'' dataset (1652 frames) achieved estimation accuracies of 99.6\% for viewpoints and 96.2\% for number of poses.
\end{abstract}

\section{Introduction}
\label{sec_intro}
The research reported in this article is motivated by the application scenario --- for example, in disaster response, where an unmanned aerial vehicle (UAV) is required to recognize the actions of a human subject, then take responsive actions. The application scenario invites the following challenges: (i) Before a UAV can even begin recognizing human actions, the UAV will first have to compute how to orientate itself towards the human subject. (ii) Many UAVs are equipped with only one monocular camera; hence, additional data  provided by stereoscopic, infrared and more advanced cameras is unavailable. (iii) Recognizing human actions from videos captured from a stationary platform is already a challenging task, owing to the articulated structure and range of poses of the human body. (iv) The difficulty of recognition is compounded by the quality of videos which include perspective distortion, occlusion and motion blur. 

We assume the UAV is in hovering flight, having a human subject within its field of view\footnote{https://asankagp.github.io/aerialgaitdataset/}. We estimate the gait sequence and movement trajectory of a person from a video captured by a UAV. Our solution consists of the following modules:  

The \emph{perspective correction module} compensates for perspective distortion in aerial videos using a projective transformation technique similar to \cite{orrite04shape,richter11perspectives,rogez14exploiting}. Instead of one homography matrix, pre-annotated homography matrices are used for different levels of distortion caused by different camera elevation angles.

The \emph{segmentation and feature extraction module} performs segmentation, using Histograms of Oriented Gradients (HOG) \cite{dalad05histograms} or alternatively, Convolutional Neural Network (CNN) \cite{lecun98gradient} features as shape descriptors. The model-free approach is combined with silhouette-based shape matching (in the case of HOG) or RGB-based shape matching (in the case of CNN) for efficient processing.

The \emph{pose estimation module} uses a dynamic classifier selection architecture inspired by \cite{woods97combination,ko08dynamic}. A total of 64 classes are defined for the combinations of 8 poses/gaits and 8 viewpoints in a human gait cycle. These views include front, back, side and diagonal views. Instead of performing 64-class classification, our dynamic classifier leverages the temporal relationships between poses and viewpoints, and performs significantly more reliable 4-class classifications instead. 
  
The \emph{trajectory estimation module} estimates the trajectory of the human subject by reconstructing the pose sequence using 3-D skeletons and localizing them with respect to the initial pose and viewpoint. The reconstructed poses provide the approximate shape of the ground-truth walking path.

The contribution of the paper is twofold.
\begin{itemize}
  \item \textbf{A classifier architecture for efficiently and robustly estimating human gaits in monocular aerial videos.} By exploiting the temporal relationships between poses and viewpoints, our method can limit the wrongly estimated viewpoints to the adjacent viewpoints of the ground truth. The loss of limb and joint details in images usually leads to a left-right ambiguity issue, especially in front and back views \cite{agarwal06recovering}. However, our pose estimation solves this problem by taking into account possible temporal transitions between states. The dynamic classifier architecture presented in this work does not execute all the classifiers in the pool to make a decision. Instead, only the relevant classifier is selected based on the state transition graphs. This is a significant difference compared to similar architectures in the literature \cite{woods97combination,kuncheva01decision,tulyakov08review}. Experimental results confirm the proposed dynamic classifier is suitable for gait estimation in both ground and aerial videos.
  \item \textbf{The creation of a training dataset from an aerial platform for gait recognition.} This dataset accounts for the natural twists and self-occlusions of a turning human body and minimizes the false positives caused by minor variations in heading.
\end{itemize}

The rest of this article is organized as follows. Section~\ref{sec_related_work} discusses closely related work on perspective correction, pose estimation and their applications to UAV-based scenarios. Section~\ref{sec_methodology} describes our solution. Section~\ref{sec_experiments} reports experimental results. Discussion of issues and potential improvements is presented in Section~\ref{sec_discussion}. Section~\ref{sec_conclusion} concludes.

\section{Related Work}
\label{sec_related_work}

This work is as an extension to the approach described in \cite{perera18human}. Compared to \cite{perera18human}, here, we propose perspective correction for reducing errors of the dynamic classifier. Compensation for the perspective distortion is analyzed for videos captured at different heights/angles. Further, we combine transfer learning and dynamic classifiers to perform CNN-based classification. The dynamic classifiers are evaluated for HOG and CNN features. The performance measures are calculated with respect to the ground truth of test videos. The dataset used for this work will be publicly available. The study reported in \cite{perera18human} has been performed only with HOG features of silhouettes, and the ground truth pose information has not been considered for performance analysis.

The problem of recognizing human pose in statically captured videos has been studied extensively in recent literature \cite{wang10review}. Here, we discuss some closely related work. 

\subsection{Perspective correction}

Perspective distortion needs to be corrected for processing that is robust to distance changes. Projective transformation, or homography, is an established approach for correcting perspective distortion \cite{hartley03multiple}, but this traditional approach requires the \emph{vanishing point} to be manually specified. Rogez et al.~\cite{rogez06viewpoint,rogez14exploiting} used \emph{vertical scene lines} to estimate the vanishing point and localize the reconstructed poses based on the vanishing point, but their approach still requires manual determination of which lines are vertical. Our homography step is similar to Rogez et al.'s, except we determine the vanishing point based on the altitude and angle of the camera. Moreover, Rogez et al. conducted their study on statically captured videos, while we use video captured from a UAV.

\subsection{Pose estimation by classification}

Dynamic classifier selection (DCS) was originally proposed by Woods et al.~\cite{woods97combination}, and is based on the \emph{local accuracy estimation} of each individual classifier. Their approach selects an individual classifier which is most likely correct for a given sample. The final decision is made only by the selected classifier. Kuncheva \cite{kuncheva02switching} proposed a classifier selection and fusion method, but the experimental results show that DCS is the best performer while Kuncheva's is the second best, provided the classifiers have the same structure and training protocol. Ko et al.~\cite{ko08dynamic} developed a relatively similar classifier by integrating a majority voting system. Our classifier follows the DCS principles, but the best individual classifier is selected without executing the entire ensemble of classifiers. 

Gaits estimated using direct classification \cite{wang10review} do not respect any temporal order \cite{xue10infrared,collins02silhouette}. A better approach is to consider the temporal relationship between poses, using techniques such as the ratio of the number of pixels in the intersection to the number of pixels in the union of two silhouette frames \cite{sarkar05humanid}, dynamic time warping \cite{veeraraghavan05matching}, lower limb joint angles \cite{zeng14model} and frequency analysis of spatio-temporal gait signatures \cite{boulgouris05gait}, to mention a few. Furthermore, general approaches to spatio-temporal action recognition can be found in \cite{sheikh05exploring,rao02action,rapantzikos11spatiotemporal,chen16action}. The temporal order of our reconstructed poses is based on the state transition model of poses and viewpoints. 

Recent human pose estimation research has shown significant performance improvements by incorporating deep learning techniques \cite{weibo17survey}. The state of the art in human pose estimation has  adopted convolutional neural networks as the main building block \cite{wei16convolutional,newell16stacked,rogez17lcr,pishchulin16deepcut}. Deep learning models have also been adopted for human trajectory estimation in a range of settings. Some notable extensions are Recurrent Neural Networks (RNN) \cite{rajiv16applying}, Behaviour-CNN \cite{yi16pedestrian}, unsupervised feature learning for classification and regression \cite{tharindu17soft}, and deep recurrent Long Short Term Memories (LSTMs) \cite{labbaci17deep}, to mention a few.

Left-right ambiguity is an inherent problem in sil\--houette-based pose estimation. Some strategies have been proposed in relation to depth information. Shotton et al.~\cite{shotton11realtime} trained classifiers to learn subtle visual cues from silhouettes to resolve the left-right ambiguity. In \cite{zhao15strategy}, the depth of each pixel in the silhouette was used to expand the 2D shape context into 3D space. Approaches have also been proposed for silhouettes in RGB images. Sigal et al.~\cite{sigal06measure} proposed switching the left-right limbs of their graphical model to fit the silhouette with the smallest error. Some notable studies incorporated temporal information to infer the correct views of silhouettes. Both Discrete Cosine Transform (DCT) temporal prior in \cite{huang17towards} and sequential clonal selection algorithm (CSA) in \cite{li14generative} handled this issue using the temporal continuity in images. Similarly, Lan et al.~\cite{lan05beyond} selected the left vs. right configuration that is most consistent with the previous frame. Our proposed algorithm also uses temporal information between images but unlike the above methods, the transitions are determined based-on simple state transition graphs.

In recent literature, transfer learning approaches have been used effectively for human motion analysis problems. In a transfer learning setting, the first selected layers of a base network are used to develop the target network \cite{yosinki14how}. Notable work was done by Chaturvedi et al. \cite{chaturvedi15deep}, that employed deep transfer learning (DTL) to analyse the trajectories of basketball players using time-delayed Gaussian networks. Mart\'in-F\'elez et al.~\cite{Martin-Felez12gait} developed a system that learned gait features independently of the identity of people by applying transfer learning on a bipartite ranking model. A transfer learning approach similar to the one described in this paper can be found in \cite{farrajota16deep}, which introduced a framework for pose and gait estimation of elderly people. Their transfer learning was based on the Alexnet model \cite{alex12imagenet}, followed by a Siamese network to compare faces and upper/full bodies. Some related work focused on recognizing human actions across changes in the observer viewpoint \cite{rahmani17learning,farhadi08learning}. Rahmani et al.~\cite{rahmani17learning} proposed a Robust Non-Linear Knowledge Transfer Model (R-NKTM) for action recognition from novel viewpoints. A similar problem was addressed by Farhadi et al. in \cite{farhadi08learning} by training a discriminative appearance model. The authors used Maximum Margin Clustering to construct split-based features in the source view, then trained  a classifier by transferring the splits in the source view to the target view.

Using UAV imagery for human pose or gait estimation is challenging due to platform mobility and susceptibility to wind gusts. UAVs are deployed in situations where it would be beneficial to interpret human movement, particularly in search and rescue applications \cite{andriluka10vision}, human-machine interface systems \cite{naseer13followme} and surveillance systems \cite{lim15monocular,aguilar17pedestrian}. When employed in surveillance or search and rescue the movement of human subjects and their trajectory are vital information. Trajectory can be used for semantic analysis of human activities and prediction of future locations from video sequences \cite{lao09automatic}. Our vision-based trajectory estimation is relatively similar to \cite{lim15monocular} in terms of visual sensing.

\subsection{UAV-based applications}

Utilizing UAVs in human tracking and action recognition missions is a relatively new topic. Human detection methods from aerial videos have been suggested in relation to search and rescue missions \cite{rudol08human,andriluka10vision}. The primary focus of these studies was to identify humans lying or sitting on the ground. Al-Naji et al.~\cite{al-naji17remote} used a hovering UAV to detect the vital signs of a human subject from the head and neck areas. Some studies focused on human identity recognition in low-resolution aerial videos. Oreifej et al.~\cite{oreifej10human} presented an algorithm relying on a weighted voter-candidate formulation. The algorithm detects targets by analyzing the ``blobs'' of candidates against voters and addresses the need for human blob detection and tracking. Yeh et al.~\cite{yeh16fast} proposed a relatively similar blob matching approach using an adaptive reference set of previously identified people. A system developed for UAV onboard gesture recognition was proposed by Monajjemi et al. in \cite{monajjemi15uav}. The system identified periodic movements of waving hands from other periodic movements like walking and running in an outdoor environment. Our experimental set-up is most similar to Monajjemi et al.'s. A crowd detection and localization approach using one UAV and a number of unmanned guided vehicles (UGVs) was presented in \cite{minaeian16vision}. In contrast, our study uses a simpler configuration, but performs robustly on aerial video.

\begin{center}
\begin{table}[th]
\caption{\label{tbl:nomenclature}Nomenclature}
\begin{tabularx}{\linewidth}{lX}
$\mat{H}$ & Homography matrix.\\
$\phi$, $\theta$ & Elevation angle, azimuth angle.\\
$\mathbb{R}$, $\mathbb{V}$, $\mathbb{P}$ & Real number space, vector space, projective space.\\
$V_i$ & The $i$th viewpoint, $i=1,\ldots,8$.\\
$P_j$ & The $j$th pose, $j=1,\ldots,8$.\\
$S$   & Training sample set.\\
$K$   & Number of classes in a training set.\\
$K'$  & Predicted class.\\
$\mat{M}$   & An ECOC coding matrix.\\
$C_{64}$    & The 64-class classifier invoked in the initialization stage.\\
$C_4(P,V)$  & The 4-class classifier associated with pose $P$ and viewpoint $V$.
\end{tabularx}
\end{table}
\end{center}

\section{Methodology}\label{sec_methodology}

\begin{figure*}[ht!]
\centering
\includegraphics[width=\textwidth]{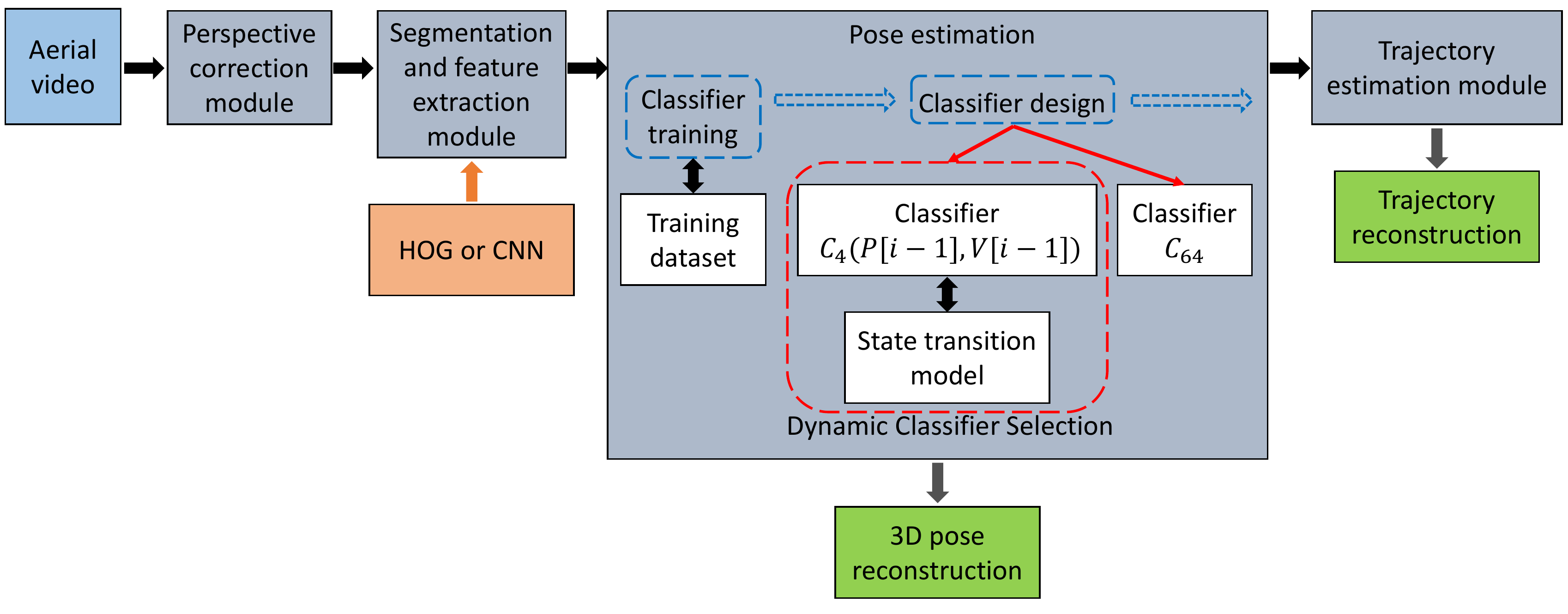}
\caption{The block diagram of the entire process. The perspective correction, segmentation and feature extraction, pose estimation: classifier training, pose estimation: classifier design, and trajectory estimation modules of the block diagram are explained from Sects.~\ref{sec_proj_trans} to \ref{sec_traj} in order.}
\label{fig_entire_process}
\end{figure*}

This section provides details of the perspective correction, segmentation and feature extraction, pose estimation and trajectory estimation modules. The block diagram of the entire process is given in Fig.~\ref{fig_entire_process}. See Table \ref{tbl:nomenclature} for the nomenclature used in this article.

\begin{figure*}[t]
\centering
\includegraphics[width=0.6\textwidth]{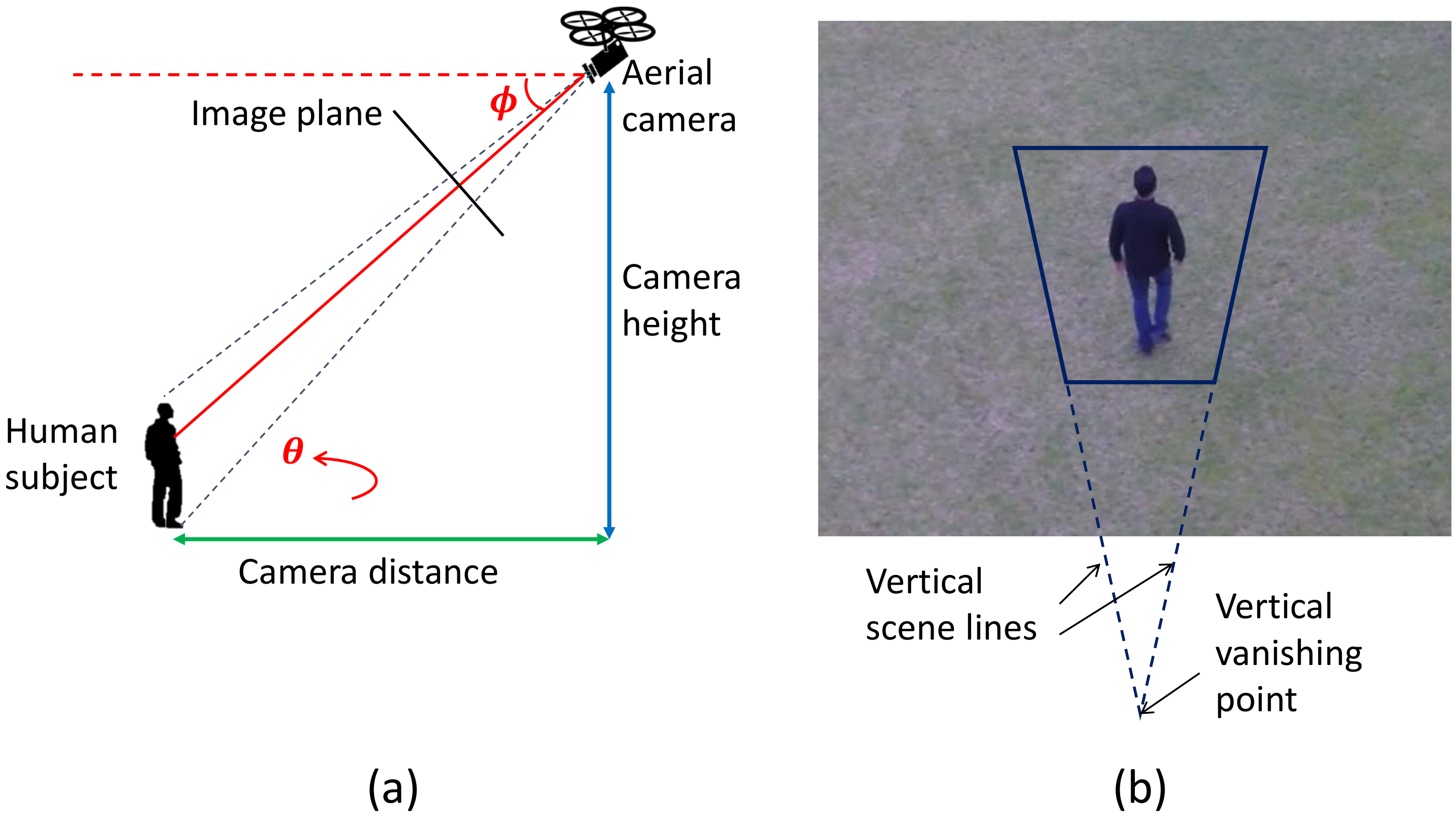}
\caption{(a) The UAV hovers at a known camera height and angle. In the horizontal coordinate system, $\phi\in[0, \pi/2]$ is the elevation/tilt angle, whereas $\theta\in[0, 2\pi)$ is the azimuth/pan angle. The azimuth angle is calculated in the radial direction between the heading direction of the human subject and the camera center axis on the horizontal plane. The observable space relative to the human observer is called the \emph{viewing hemisphere}. (b) An input image with vertical scene lines. The scene lines are manually constructed according to the elevation angle. The blue box in the middle is the area of interest for homography in the vertical plane. The vanishing point is the point where parallel scene lines would meet each other on the image plane.}
\label{fig_perspective}
\end{figure*}

\subsection{Perspective correction}
\label{sec_proj_trans}

The relative orientation between the human subject on the ground and the camera in the sky is captured in a \emph{horizontal coordinate system}, with coordinates $\phi$ and $\theta$ (see Fig.~\ref{fig_perspective}(a)). The camera viewpoint can take any $(\phi,\theta)$ pair depending on the UAV position, where $\phi\in[0, \pi/2]$ and $\theta\in[0, 2\pi)$. A major problem with aerial photography is vertical perspective distortion, which occurs when $\phi>0$, and worsens as $\phi$ gets larger. At low altitude the distorted human shape tends to have a large head and shoulders and small feet (see Fig.~\ref{fig_perspective}(b)). When $\phi=90^\circ$, perspective distortion cannot be corrected. For $60^\circ\leq\phi<90^\circ$, the captured images have a severely distorted perspective that is difficult to accurately compensate. Therefore in this study, we limit the maximum $\phi$ to $60^\circ$.

Perspective correction is done by mapping the distorted image plane (see Fig.~\ref{fig_perspective}(b)) to the undistorted vertical plane through \emph{homography}. Segments on the undistorted vertical plane then enable the matching of test and training images.

A homography is a mapping from a projective space to itself. A \emph{projective space} is an extension of Euclidean space in which two lines always meet at a point, and a point in the projective space is called a \emph{homogeneous point}. Given an image, for every homogeneous point on the image plane, $\vec{x}$, there exists a homography matrix $\mat{H}$ \cite[Section 3.1]{smith04invitation} that maps it to a homogeneous point, $\vec{x}'$, on the undistorted vertical plane, i.e.,
\begin{equation*}
\vec{x}' = \mat{H}\vec{x}.
\end{equation*}
The matrix $\mat{H}$ depends on the elevation angle $\phi$. Instead of calculating $\mat{H}$ for each frame, we calculate it offline for each of the following values of $\phi$: 
\begin{itemize}
  \item $\arctan(10/30)=18.4^\circ$,
  \item $\arctan(20/30)=33.7^\circ$,
  \item $\arctan(30/30)=45.0^\circ$,
  \item $\arctan(40/30)=53.1^\circ$;
\end{itemize}
and manually pre-annotate videos of the same elevation angle with the corresponding $\mat{H}$. To calculate $\mat{H}$, we manually select four points in a sample video frame to (i) delineate the area of interest and (ii) generate the vertical scene lines, as shown in Fig.~\ref{fig_perspective}(b). The vertical scene lines define the homography matrix $\mat{H}$.

\subsection{Segmentation and feature extraction}\label{sec_segmentation}

After perspective correction, the human silhouette is segmented. The size of the silhouette in the image plane varies depending on the direct distance between the camera and the human subject. Perspective correction alone cannot address this scaling issue. Thus, the test silhouette is scaled up or down to match the scale of the training images. Prior to feature extraction, we use the online video annotation tool VATIC~\cite{vondrick13efficiently} to annotate the test videos. Two types of features are used, resulting in two variants of our scheme: HOG-based and CNN-based.

\subsubsection{Feature extraction using HOG}

For each frame, the RGB image is converted into a binary image and its bounding box area is segmented. Noise is removed using a Gaussian filter and small objects containing fewer than a threshold number of pixels are also removed. The remaining blob or blobs are considered to represent the human silhouette. Currently, the denoising parameters and segmentation parameters are customized for each video clip to obtain the best possible silhouette, so they are subject to improvements.

For feature extraction, the image window is divided into small spatial regions called ``HOG cells'' \cite{dalad05histograms}. The weighted gradients in a HOG cell form a 1-D histogram which represents the orientation of the edge lines. The feature vector is formed from the HOG blocks, each of which represents a group of HOG cells.

\begin{figure*}[t]
  \begin{center}
  \includegraphics[width=0.8\linewidth]{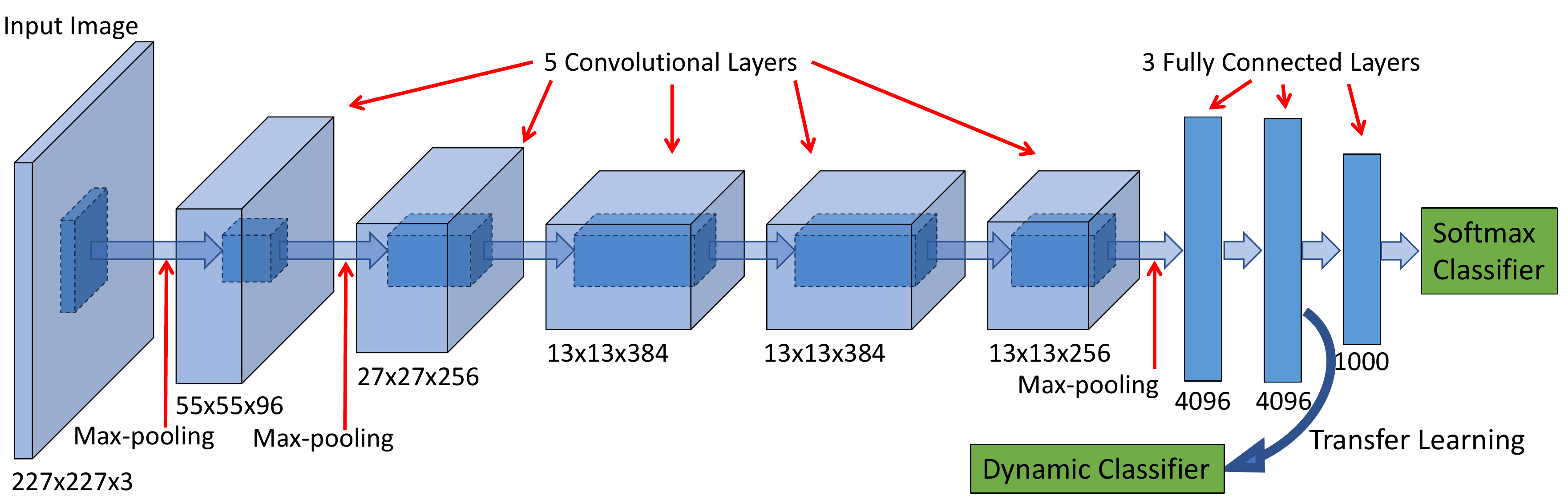}
  \caption{The architecture of the pre-trained AlexNet deep convolutional neural network. The feature vector is extracted before the last fully connected layer and used to train the dynamic classifier.}
  \label{fig_alexnet}
  \end{center}
\end{figure*}

\subsubsection{Feature extraction using CNN}\label{sec_cnn_feature}

For each frame, the RGB image is cropped to meet the size requirement of the deep CNN AlexNet~\cite{alex12imagenet}. The AlexNet is a 11-layer network including 5 convolutional layers, 3 fully connected layers and 3 max pooling layers (see Fig.~\ref{fig_alexnet}). The early convolutional layers have small receptive field sizes for learning low-level features, and later layers have larger field sizes for learning higher-level features. AlexNet has been pre-trained on 1.2 million ImageNet \cite{deng09imagenet} images of 1000 classes, and showed the best performance in the ImageNet Large Scale Visual Recognition Challenge in 2012 \cite{alex12imagenet}. Some classes of AlexNet are trained on images of humans in different settings; therefore, this pre-trained network was selected for our work.

In a pre-trained network, the weights for the deep layers are pre-determined. Instead of re-training AlexNet with our comparatively small dataset, we apply \emph{transfer learning} in the standard way. We take the 4096-dimensional vector right before the last fully-connected layer of AlexNet as the feature vector (see Fig.~\ref{fig_alexnet}). We then use the feature vectors to train an SVM classifier (as described in Section~\ref{sec_classifier_design}).

\subsection{Pose estimation: classifier training}\label{sec_training}

\begin{figure}[ht]
\begin{center}
\includegraphics[width=\linewidth]{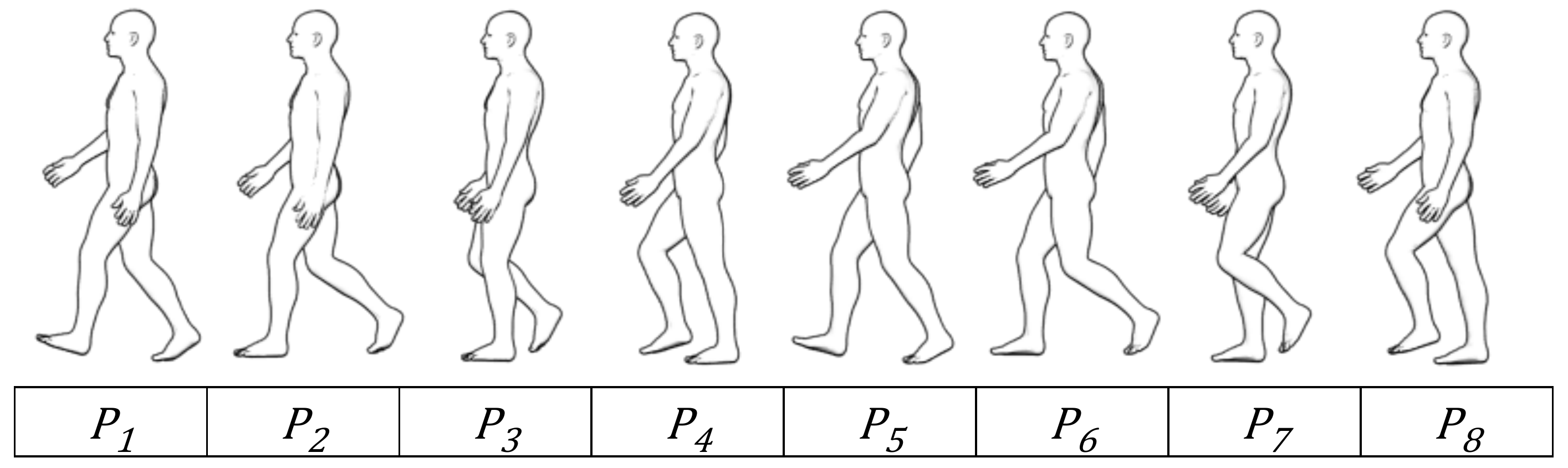}
\caption{A representation of the human gait cycle in 8 sub-steps: $P_1$ -- both legs touch the ground; $(P_2-P_4)$ -- swinging the right leg; $P_5$ -- both legs touch the ground; $(P_6-P_8)$ -- swinging the left leg.} 
\label{fig_gait_cycle}
\end{center}
\end{figure}

\begin{figure}[ht]
\begin{center}
\includegraphics[scale=0.4]{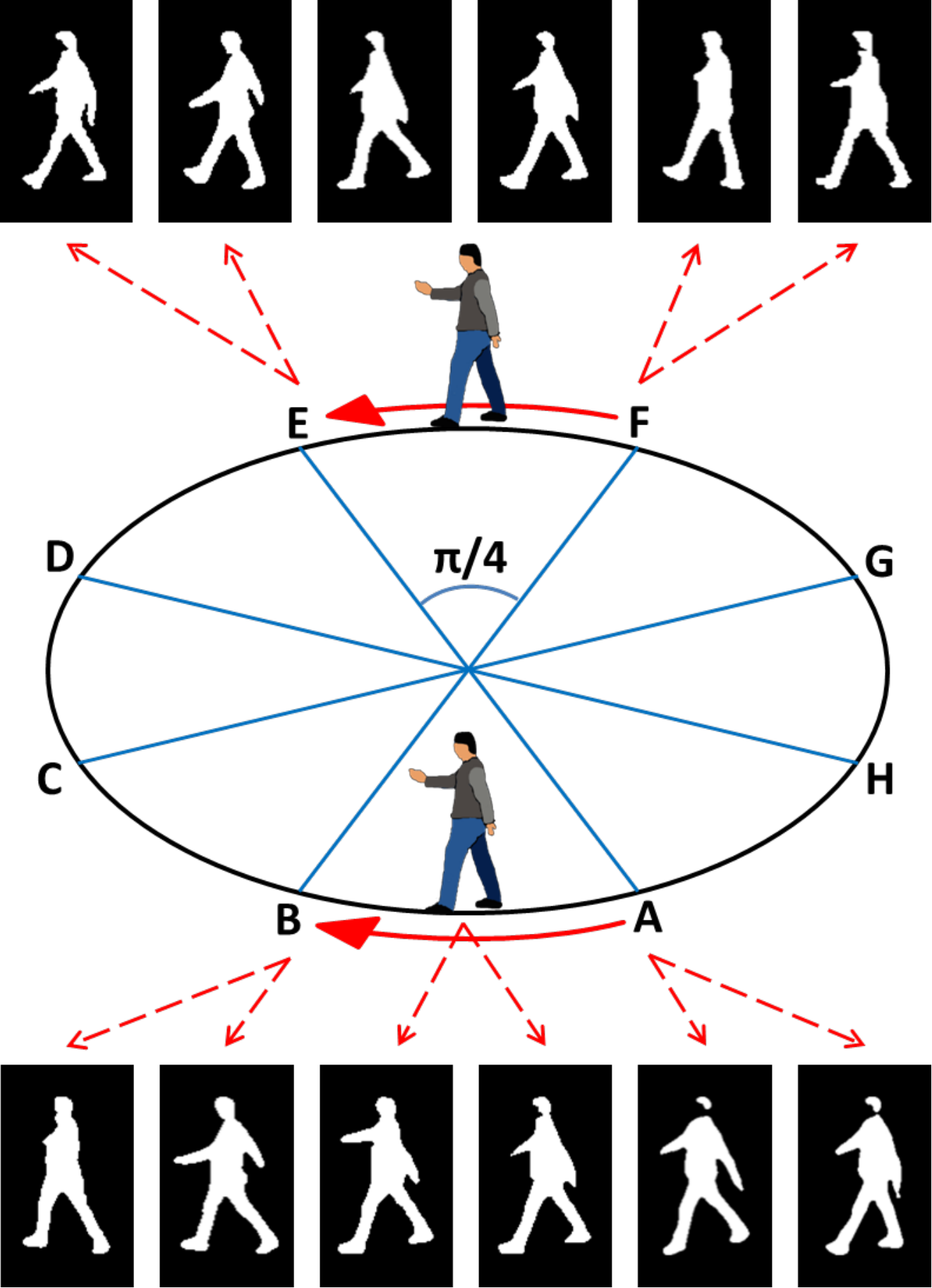}
\caption{The silhouettes in the top and bottom rows correspond to the first sub-step of human gait cycle shown in Fig.~\ref{fig_gait_cycle}. For that particular pose there are eight possible viewpoints. This figure shows how the training images are selected for the shown pose and viewpoint. The broken arrows point to the selected silhouettes within the allowed orientation error margin. The starting point of the arrow roughly indicates the silhouette location on the circle. The silhouettes are independently selected from different image sequences captured under different lighting and noise conditions.} 
\label{fig_training_dataset}
\end{center}
\end{figure}

\noindent The training dataset is created to identify the eight sub-steps~\cite{whittle14gait} of the human gait cycle (see Fig.~\ref{fig_gait_cycle}). Each sub-step (or pose) has viewpoints from eight radial directions (azimuth angles that are $45^\circ$ apart), giving rise to $8\times8=64$ pose-viewpoint pairs. The finite number of elevation-azimuth angle pairs are equivalent to the discretized viewing hemisphere described in~\cite{rogez14exploiting,rogez06viewpoint,rosales06combining}. 

We create two training datasets using silhouettes and color images. The silhouette dataset has 1017 images across 64 classes. The color image dataset contains images from our field videos, the MoBo Aligned dataset~\cite{rogez12fast} and the HumanEva dataset~\cite{sigal09humaneva}. To create this dataset, we collected images with varying perspective distortion for each class. The field images recorded from moving subjects have varying perspective distortions because the subjects walked in a circle (see the first three images in Fig.~\ref{fig_color_tr_images}). The original MoBo images \cite{gross01cmu} were recorded on a treadmill using fixed cameras. We changed the backgrounds of MoBo Aligned images and manually added some perspective distortion to the images (see the second three images in Fig.~\ref{fig_color_tr_images}) and no modifications were done to HumanEva images (the last two images in Fig.~\ref{fig_color_tr_images}). We manually selected 4 points on the MoBo Aligned images and applied perspective transformation to get a bird’s eye view of them. We collected field images of 3 subjects, MoBo Aligned images of 15 subjects and HumanEva images of 2 subjects. The color dataset contains 8111 color images across 64 classes.

We used the entire training dataset (silhouettes or color images) to train the classifiers. The testing has been done using five selected videos (silhouettes or color images): a video each from CMU Motion of Body (MoBo) \cite{gross01cmu} and HumanEva2 \cite{sigal09humaneva} datasets (see Sect.~\ref{sec:exp:pub} for more details) and three aerial videos (see Sect.~\ref{sec:exp:uav} for more details). The annotated training and testing data will be publicly released in 2018.

\begin{figure}[t]
\begin{center}
\includegraphics[scale=0.3]{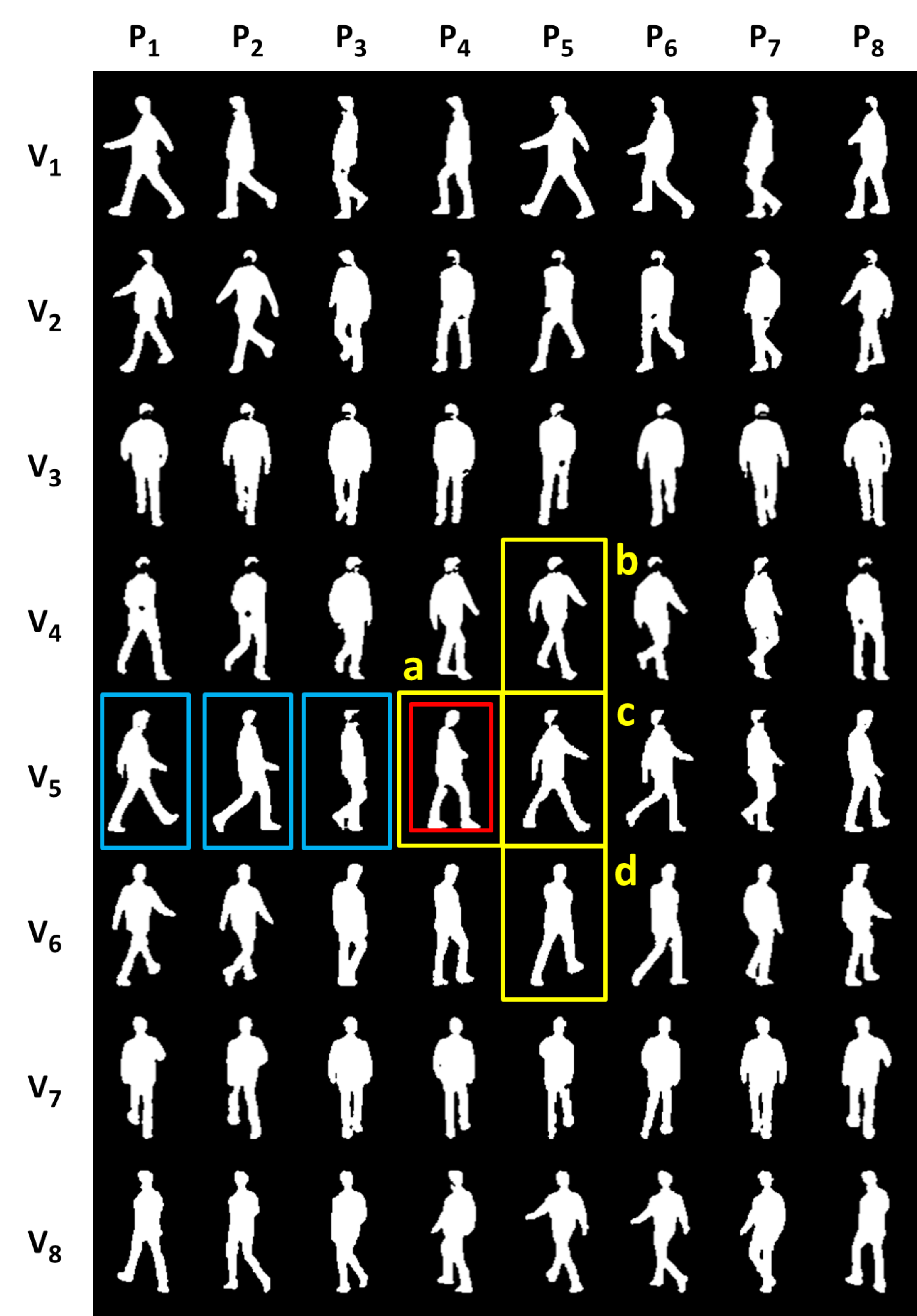}
\caption{The training dataset consists of 64 classes of pose-viewpoint pairs. The rows and columns represent the poses and the viewpoints respectively. Each silhouette in the figure is a random image from the pose-viewpoint subset it belongs. The initialization is performed using the classifier trained with the complete dataset (64 classes), namely $C_{64}$. Blue- and red-border windows show four consecutive frames (from left to right) initialized by this classifier. Once initialized, the pose and the viewpoint of the most recently initialized image (red-border window) are used to select the next classifier. In this example, the training image subsets of the next classifier are shown in yellow-border windows.} 
\label{fig_tr_classes}
\end{center}
\end{figure}

\begin{figure*}[t]
\begin{center}
\includegraphics[width=0.8\textwidth]{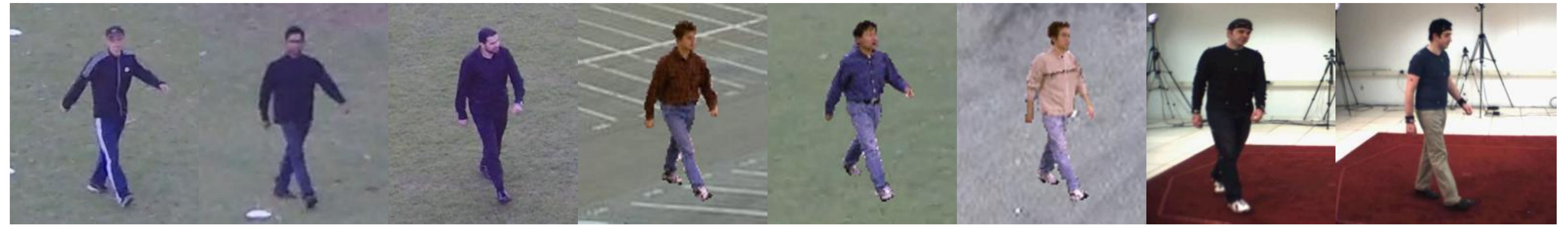}
\caption{Some sample images from the color training dataset.} 
\label{fig_color_tr_images}
\end{center}
\end{figure*}

Figures \ref{fig_training_dataset}--\ref{fig_tr_classes} show only the silhouette images, but the same technique is followed to create the color dataset. An example of training data collection is shown in Fig.~\ref{fig_training_dataset}. The silhouettes in the figure correspond to the first sub-step of the human gait cycle shown in Fig.~\ref{fig_gait_cycle}, namely $P_1$. The training data are collected at a camera distance of 30m and camera height of 10m (i.e., $\phi=18.4^\circ$), while the human subject walks on a marked circle of radius 5m in clockwise and anticlockwise directions. When walking from $A$ to $B$ on the circle, the orientation gradually changes up to $45^\circ$ with respect to the orientation at $A$. In the training dataset, the images corresponding to the walk from $A$ to $B$ are considered as walking a straight line from $A$ to $B$. This assumption can introduce a maximum of $22.5^\circ$ orientation error. This error can be reduced by selecting more viewpoints (in other words selecting a viewpoint separation angle of less than $45^\circ$). However, we limit this study to eight viewpoints for  simplicity and efficiency. The images captured at locations $A$, $B$, $E$ and $F$ on the circle have the maximum orientation error. Only the images captured at the mid-points of chords $AB$ and $EF$ represent the true orientation. The training images are selected in order to cover all of the possible heading directions within the accepted error margin ($\pm22.5^\circ$). The same procedure is followed to create the other 63 pose-viewpoint pairs as well. When the training dataset is used to estimate the poses in a test video of a person walking in a circle, the reconstructed path will not be a perfect circle (even assuming zero estimation errors) but a polygon. The reason for this polygon shape is the orientation angle error associated with each viewpoint. 

One advantage of the training data collection method above is it accounts for the natural twists and self-occlusions of walking better than collecting data only from walking straight. For example, the images captured at points $A$ and $E$ in Fig.~\ref{fig_training_dataset} have the same orientation error. However, the silhouettes of the same pose can hold differences as the person at $A$ turns to his right and the person at $E$ turns to his left on the circle. Another significant advantage is it reduces the false positives of the classifier arising from slight variations of heading. By including slightly oriented silhouettes (with respect to walking straight) in the dataset, we approximate all of the small variations in orientation to be within the range $[0^\circ, 22.5^\circ]$. This is a useful approximation when analyzing real walking patterns of human subjects because most of the time, people walk in straight lines and do not change their orientation frequently. 

The collected training data consists of 64 labels, representing eight sub-steps of the gait cycle and eight viewpoints (see Fig.~\ref{fig_tr_classes}). For each label, the training data consists of silhouettes (for HOG) and RGB images (for CNN) created under different illuminations and orientations. Some sample images of class $P_5V_4$ are shown in Fig.~\ref{fig_color_tr_images}.

\subsection{Pose estimation: classifier design}\label{sec_classifier_design}

We denote a training dataset of $n$ observations by 
\begin{equation}
S=\{(\vec{x}_1,y_1), (\vec{x}_2,y_2), \ldots, (\vec{x}_n,y_n)\}, 
\end{equation}
where $\vec{x}_i$ is the $i$th feature vector, and $y_i$ the $i$th label. Suppose $\vec{x}_i\in\mathcal{X}\subset\mathbb{R}^m$, where $m$ is the dimension of a feature vector; and $y\in\mathcal{Y}=\{1,...,K\}$, where $K$ is the number of classes. We can formulate the pose estimation problem, like most classification problems in computer vision~\cite{garcia06improving}, as a $K$-class classification problem: finding $f:\mathcal{X}\rightarrow\mathcal{Y}$ such that the classification error is minimized.

However, many real world problems are multiclass problems, $K>2$. A standard way to create multiclass classifiers such as multiclass SVM is to map a multiclass problem onto many, possibly simpler, twoclass problems \cite{garcia06improving}. A potential solution is a classifier combination method, the basic idea of which is to execute an ensemble of classifiers, and combine their outputs through a voting system, combination function, or weighting function \cite{tulyakov08review}. In such a design, although each classifier is trained with a subset of the entire training set, each iteration involves the entire ensemble of classifiers. As a more efficient solution, we propose a \emph{dynamic classifier selection} architecture, combining (i) a state transition model for the pose and viewpoint and (ii) an SVM-based error-correcting output codes (ECOC) framework~\cite{dietterich95solving} for our multiclass pose-viewpoint classification problem. Next, we will discuss this state transition model and ECOC framework in turn.

\subsubsection{Viewpoint and pose transition}
\label{subsub_viewpoint_pose_transition}

\begin{figure*}[t]
\begin{center}
\includegraphics[width=0.8\textwidth]{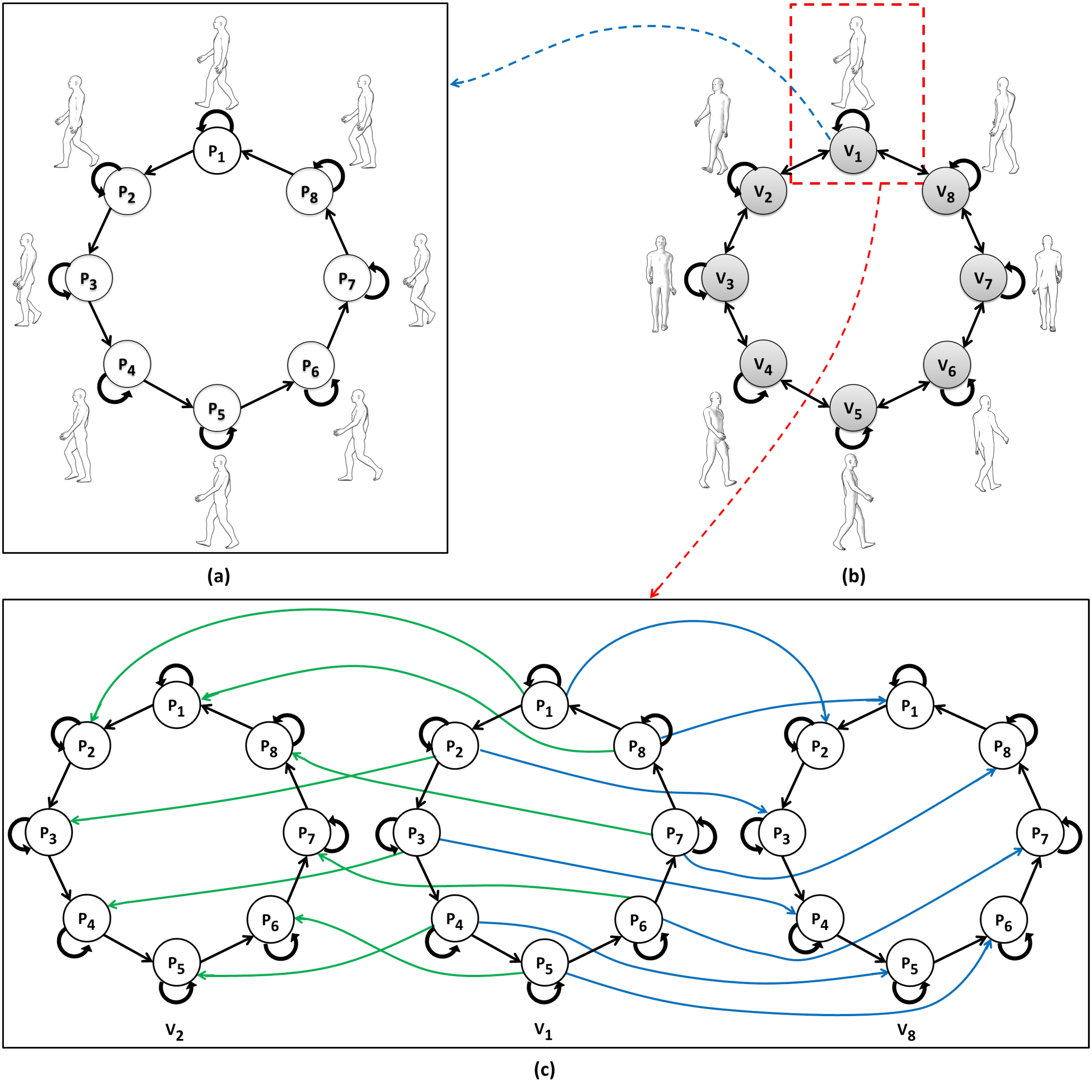}
\caption{Our state transition model for the pose and viewpoint. Here, $V_i$ represents ``viewpoint $i$'' and $P_j$ represents ``pose $j$''.  (a) The pose transition graph for viewpoint $V_1$. (b) The viewpoint transition graph, where each node corresponds to eight poses. Each viewpoint can transition to one of the immediately adjacent viewpoints, or can stay as itself. (c) The pose and viewpoint transition graphs for the selected viewpoint in (b), namely $V_1$.} 
\label{fig_stateflow}
\end{center}
\end{figure*}

We model a pose-viewpoint pair as a state, and the transition of states using a state transition graph. This graph should not be confused with a Markov chain, because we do not assign probabilities to state transitions. Our state transition model is similar to Lan et al.'s model~\cite{lan04unified}, with the differences being:
\begin{itemize}
  \item Lan et al. used 8 viewpoints and different numbers of poses per viewpoint, whereas we use 8 viewpoints and 8 poses per viewpoint;
  \item each of their side views and $45^\circ$ views is associated with 4 poses, and each of their front and back views is associated with 1 pose, resulting in a total of 26 states; whereas, our model consists of 64 states. 
\end{itemize}

Our state transition graphs (see Fig.~\ref{fig_stateflow}) are constructed based on the assumption that the human subject walks forward at a constant speed, does not take sharp turns and does not twist their body while turning. This assumption does not preclude left, right, or backward turns, as long as the turn is not abrupt, as exemplified by the yellow-border windows in Fig.~\ref{fig_tr_classes}. The state transition graphs establish a temporal relationship between the states, and dictate admissible state transitions, on which the classification outputs are based.

The admissible state transitions restrict the next classifier prediction to be one of the states the current state can transition to. Given the current pose and viewpoint, when a new image is available, the associated pose is predicted to be either the current pose (conceivably, the same pose appears in multiple consecutive frames when the video frame rate is high), or the pose in the next sub-step of the gait cycle (see Fig.~\ref{fig_gait_cycle}). When the pose changes from the current state to the next state, the viewpoint of the next pose has to be  one of the following: the same viewpoint (moving straight), $45^\circ$ clockwise from the current viewpoint (turning left), or $45^\circ$ anticlockwise from the current viewpoint (turning right). 

\subsubsection{Classification with error-correcting output codes (ECOC)}

Considering the pose-viewpoint pairs as multiple classes in the problem domain, we select the ECOC framework~\cite{dietterich95solving} for multiclass classification. The ECOC  is considered to be a powerful and popular multiclass classification technique \cite{furnkranz02round}. Good results have been reported in \cite{furnkranz02round,masulli04effectiveness,masulli00effectiveness,ghani00using} using ECOC for different multiclass classification problems.

The ECOC framework uses a set of binary classifiers to solve a multiclass classification problem. In a problem domain of $K$ classes, the ECOC framework forms $N$ binary problems (dichotomizers), where $N>K$. Dietterich et al.~\cite{dietterich95solving} represent the ECOC model of $N$ binary problems using a coding matrix $\mat{M}=[m_{k,n}]=\{-1,+1\}^{K\times N}$, where each row encodes an $N$-dimensional binary vector (a codeword), and each column is used to train a binary learner. The coding design is such that $+1$ represents a positive example of a class, whereas $-1$ represents a negative example.

We use the ternary ECOC framework proposed by Allwein et al.~\cite{allwein00reducing} that follows the steps below:
\begin{itemize}
  \item The coding matrix is defined as $\mat{M}=\{-1,0,+1\}^{K\times N}$, where $0$ tells the binary learner to ignore the corresponding class during training. 
  \item We use one-versus-one~\cite{hastie98classification} coding design which constructs $K(K-1)/2$ binary learners.
  \item The selected decoding scheme is loss-based decoding \cite{allwein00reducing}, and the binary learner is an SVM learner.
  \item In the classification stage, when an input $x$ is available, the vector of predictions $\vec{f}(x)=[f_1(x)\;\cdots\;f_N(x)]$ is formed from the predicted outputs of the $N$ classifiers. 
  \item The predicted class is the class that minimizes some loss function $L$ \cite[Equation (5)]{allwein00reducing}:
  \begin{equation}
  K' = \argminD{k\in\{1,\ldots,K\}} \sum_{n=1}^{N} L(m_{k,n},f_n(x)).
  \end{equation}  
\end{itemize}

\subsubsection{Classifier combination by dynamic classifier selection (DCS)}

Our DCS architecture consists of a single 64-class SVM classifier denoted $C_{64}$, and 64 4-class SVM classifiers denoted $C_4(P_i,V_j)$, $i,j\in\{1,\ldots,8\}$. The classifier $C_4(P_i,V_j)$ is associated with pose $P_i$ and viewpoint $V_j$, and is trained to recognize the set of four classes:
\begin{equation}
\{(P_i,V_j), (P_{i\boxplus1},V_j), (P_{i\boxplus1},V_{j\boxminus1}), (P_{i\boxplus1},V_{j\boxplus1})\},
\end{equation}
where $i,j\in\{1,\ldots,8\}$ and the operators $\boxplus,\boxminus$ are defined as follows:
\begin{gather}
i \boxplus  j = (i + j + 1) \mod 8 - 1, \\
i \boxminus j = (i - j - 1) \mod 8 + 1.
\end{gather}
For example, the classifier $C_4(P_4,V_5)$ is trained to recognize the four classes labeled $a$, $b$, $c$ and $d$ in Fig.~\ref{fig_tr_classes}.

As depicted in Algorithm~\ref{alg_flowchart}, our classification process works in two stages: (i) the initialization stage and (ii) the DCS stage. In the initialization stage, the first $q$ video frames are classified using classifier $C_{64}$. The DCS stage starts with the $(q+1)$th video frame. In this stage, each frame is classified with a classifier chosen based on the class label predicted by the previous iteration.

\begin{algorithm}[]
\SetAlgoLined
\textbf{input}: Labeled test images\;
 $i=1$\;
 \While{the $i^{th}$ video frame}{
  \eIf{$i\leq q$}{
   Initialization stage: apply classifier $C_{64}$\;
   }{
   DCS stage: apply classifier $C_4(P[i-1],V[i-1])$\;
  }
  $P[i] \leftarrow$ Predicted pose\;
  $V[i] \leftarrow$ Predicted view\;
  $i=i+1$\;
 }
 \caption{Algorithm for pose-viewpoint classification by dynamic classifier selection. Prior to the workflow, all classifiers should have been trained. Upon entering the workflow, classifier $C_{64}$ is used to classify the first $q$ video frames --- a higher value of $q$ implies higher conservativeness. Based on the output of $C_{64}$ on the $q$th frame, a classifier $C_4(P[q],V[q])$ is selected and used to classify the $(q+1)$th frame. Thereafter, $C_4(P[i-1],V[i-1])$ is used to classify the $i$th frame. Each classifier $C_4(\cdot,\cdot)$ recognizes only four classes.}
 \label{alg_flowchart}
\end{algorithm}

To elaborate, consider the example in Fig.~\ref{fig_tr_classes}. Suppose $q=4$, and the blue- and red-border windows are sample classes predicated by the classifier $C_{64}$. The red-border window highlights the class predicted for the $q$th frame. Since this class is $(P_4,V_5)$, the classifier $C_4(P_4,V_5)$ is chosen to classify the $(q+1)$th frame. The training subsets for $C_4(P_4,V_5)$ are highlighted with the yellow-border windows $a$, $b$, $c$ and $d$. A training subset refers to the images corresponding to a single class \cite{ko08dynamic}.

The most significant difference between the classifier architecture presented here and architectures in the recent literature \cite{woods97combination,kuncheva01decision,tulyakov08review} is that this architecture does not execute all of the classifiers to make a decision. Instead, only the relevant classifier is selected for every next image. The relevance of the classifier is determined by its training subsets, and the training subsets are selected based on the state transition graphs. 

In Algorithm~\ref{alg_flowchart}, the most resource-demanding component is ECOC SVM classification. The time and space complexities of this component are $\mathcal{O}(\nSV)$ and $\mathcal{O}(\nSV m)$ respectively, where $\nSV$ is the number of support vectors, and $m$ is the number of features. 

When using HOG features, the SVM model was trained using a one-versus-one coding design, which involves $K(K-1)/2$ support vectors. Final cropped silhouettes were resized to $96\times160$ pixels. The HOG cell size was selected to be $4\times4$ resulting in a 32292-dimensional feature vector. In this case, for the 4-class dynamic classifier, $\nSV$ is 6 and $m$ is 32292. 

When using CNN features, the SVM model was trained using a one-versus-all coding design, which involves $K$ support vectors. In this case, for the 4-class dynamic classifier, $\nSV$ is 4 and $m$ is 4096.

\subsection{Trajectory estimation}\label{sec_traj}

 \begin{algorithm}[]
\SetAlgoLined
 \textbf{input}: Estimated pose and viewpoint\;
 $i=0$\;
  \textbf{Label 1}: $i=i+1$\;
  Obtain the $i^{th}$ estimated pose and viewpoint\;

  \eIf{$i=1$}{
       Initialization: calculate orientation, reconstruct 3-D pose\;
       go to \textbf{Label 1}\;
   }{
      \eIf{pose changed?}{
          \eIf{viewpoint changed?}{
              Calculate new orientation\;
              Reconstruct 3-D pose\;
              Rotate towards new orientation\;
              Translate along new orientation\;
              go to \textbf{Label 1}\;
          }{
              Reconstruct 3-D pose\;
              Translate along previous orientation\;
              go to \textbf{Label 1}\;
          }
      }{
          go to \textbf{Label 1}\;
      }
  }
 \caption{Algorithm for trajectory estimation.}
 \label{alg_localization}
\end{algorithm}

Trajectory estimation refers to estimation of the shape of path traversed by the human subject. Trajectory estimation is performed using the estimated viewpoints as inputs and thus, require the classifier to have minimal viewpoint estimation errors. The estimated trajectory is inevitably a polygonal approximation of the actual shape of the path. Various interpolation techniques could be applied to smoothen the estimated trajectory and thereby improve the approximation.

As shown in Algorithm~\ref{alg_localization}, each estimated viewpoint serves as an estimation of the walker's orientation. For each estimated orientation, a 3-D pose is reconstructed from the estimated pose. The algorithm can be thought of as primarily handling two cases:
\begin{itemize}
  \item Whenever an estimated pose is the same as the previous, the reason is assumed to be the camera's high frame rate and/or the subject's slow movement, and thus the subject is assumed to remain at the same location. 
  \item Whenever an estimated pose differs from the previous, the subject is assumed to have moved a fixed distance from the location of the previous pose. When the orientation changes by $x$ degrees, the next pose is positioned at a fixed distance from the location of the previous pose at an angle of $\pm x$ degrees ($+$ve for right turns, $-$ve for left turns). Due to the way the viewpoint angle is discretized, as explained in Section~\ref{sec_training}, $x$ is a multiple of $360^\circ/8=45^\circ$.
\end{itemize}

The trajectory estimation algorithm uses the dynamic classifier's ability to resolve the left-right ambiguities of images. Without dynamic classifier selection, the classifier $C_{64}$ can make errors between the front and back views (rows 3 and 7 in Fig.~\ref{fig_tr_classes}), as a result of self-occlusions, or loss of joint angle and limb length information after binary conversion. The time and space complexities of Algorithm~\ref{alg_localization} are both $\mathcal{O}(1)$.

\begin{figure*}[t]
\begin{center}
\includegraphics[width=0.75\textwidth]{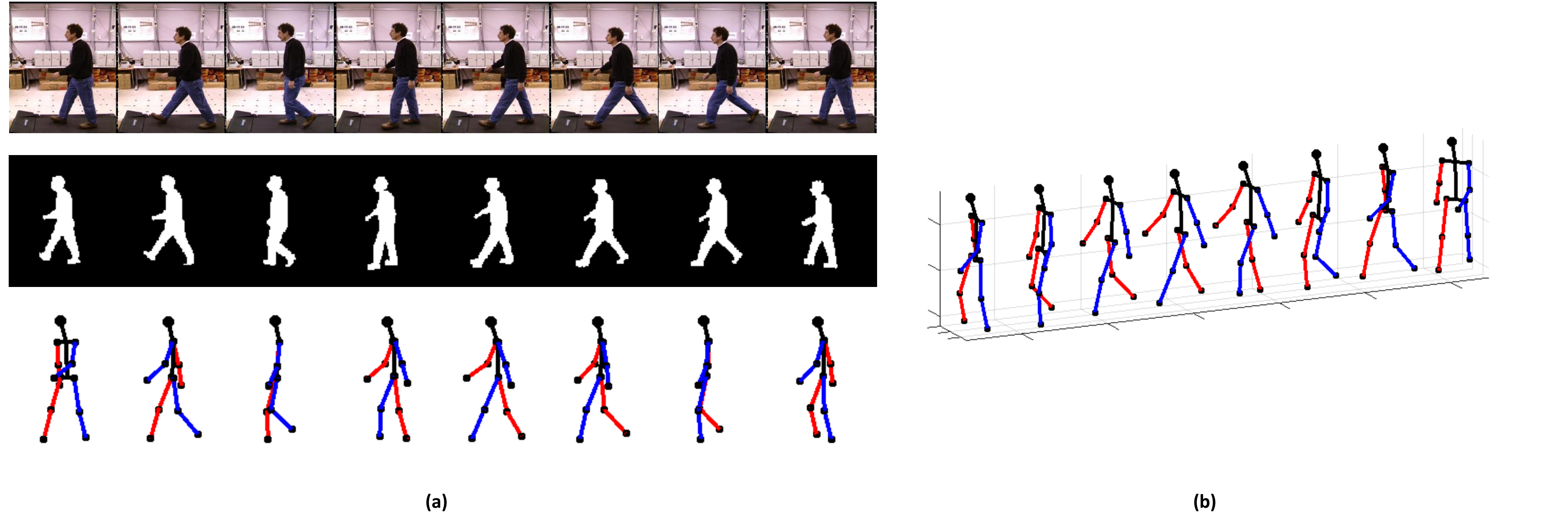}
\caption{Partial results for the CMU MoBo experiments: (a) The three rows represent the original images for subject 4071, the segmented silhouettes and the estimated poses respectively. (b) 3-D reconstruction of the estimated poses and trajectory of (a), where the rightmost skeleton corresponds to the leftmost image in (a). Red and blue colors mark the right and left sides of the body respectively.} 
\label{fig_cmu_mobo}
\end{center}
\end{figure*}

\begin{figure*}[t]
\begin{center}
\includegraphics[width=0.75\textwidth]{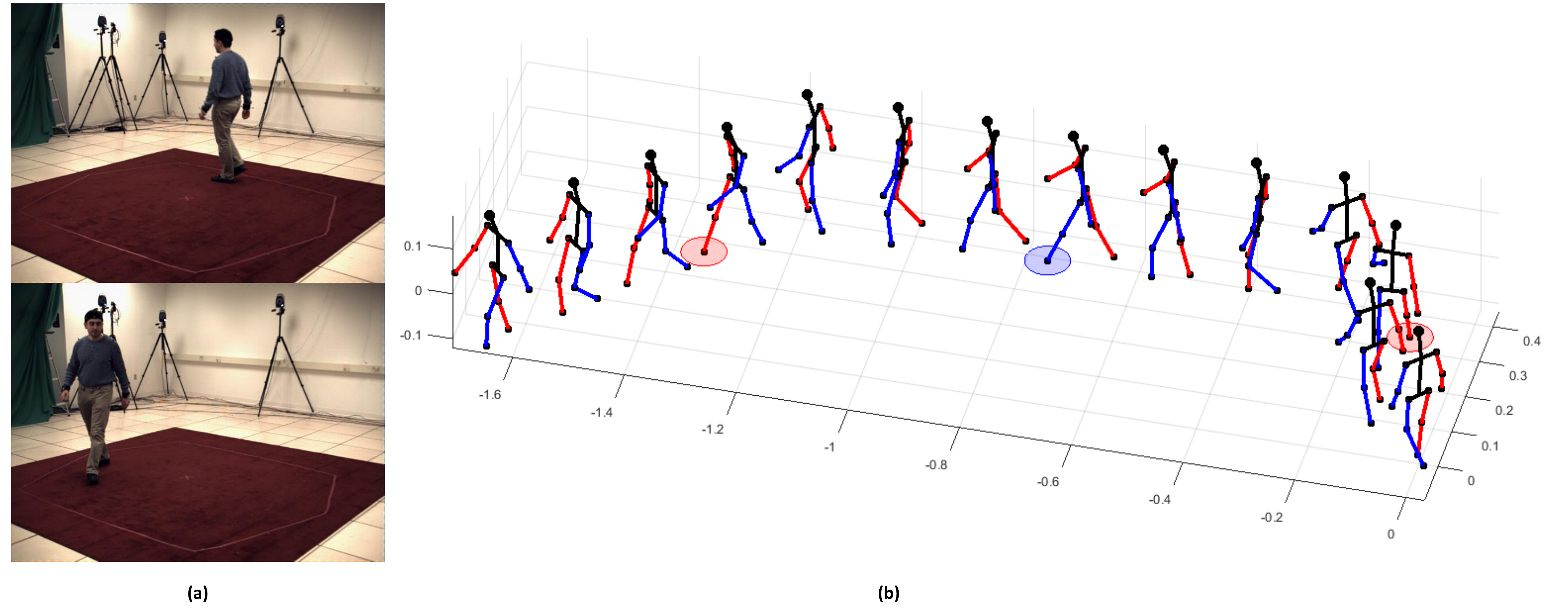}
\caption{Partial results for the HumanEva2 experiments: (a) The first and last images of the sequence for subject S2 combo C2. (b) 3-D reconstruction of the estimated poses and trajectory. Red and blue discs mark the sub-steps of the human gait cycle where both feet touched the ground --- red when the right foot was in front and blue when the left foot was in front. The 3-D poses between two red or two blue markers represent a complete gait cycle (8 poses).} 
\label{fig_humaneva2}
\end{center}
\end{figure*}

\section{Experimental results}\label{sec_experiments}

We conducted three groups of experiments, which we discuss in the subsequent subsections. Across the experiments, the scenery and walking patterns vary significantly. These experiments include view variations between front, diagonal, side and back views.

Pose/viewpoint estimation errors are expressed in terms of (i) classification errors and (ii) viewpoint and pose transitional errors (TE). We define \emph{transitional errors} as follows: 
\begin{Definition}
If the classifier prediction is different from the ground truth but is confined between the adjacent viewpoints (or poses), such predictions are considered as \emph{viewpoint transition errors} (or \emph{pose transition errors}).
\end{Definition}
\noindent For example, when the ground-truth pose transitions from $P_1$ to $P_2$, given the similarity of the poses, it is likely for the classifier to still identify ground truth $P_2$ as $P_1$ for a few frames. Likewise, the classifier is likely to misclassify $P_1$ as $P_2$ before the ground-truth transition occurs. In other words, transitional errors can delay or advance true pose/viewpoint estimation. In this example, the performance measures without transitional errors are calculated as follows: when the ground truth is $P_2$, the predicted adjacent poses ($P_1$ and $P_3$) are considered to be true predictions, and all the other estimations are considered to be incorrect. Hereafter, we use the abbreviation TE in the equations, tables and figures to refer to transitional errors. 

We now define pose/viewpoint estimation errors formally:
\begin{Definition}
  The percent \textbf{pose estimation error}, including transitional errors, is
  \begin{equation}\label{eq:e_pose_te}
  e_\text{pose, with TE} \definedas \frac{\Big|\substack{\text{\#frames with misclassified poses}}\Big|}{\text{\#frames}}\times100\%.
  \end{equation}
  %
  The percent \textbf{pose estimation error}, excluding transitional errors, is
  \begin{equation}\label{eq:e_pose_no_te}
  e_\text{pose, no TE} \definedas \frac{\Big|\substack{\text{\#frames with misclassified poses} \\ -\text{\#frames with pose TE}}\Big|}{\text{\#frames}}\times100\%.
  \end{equation}
  %
  The percent \textbf{viewpoint estimation error}, including transitional errors, is
  \begin{equation}\label{eq:e_viewpoint_te}
  e_\text{viewpoint, with TE} \definedas \frac {\Big|\substack{\text{\#frames with misclassified} \\ \text{viewpoints}}\Big|}{\text{\#frames}}\times100\%.
  \end{equation}
  %
  The percent \textbf{viewpoint estimation error}, excluding transitional errors, is
  \begin{equation}\label{eq:e_viewpoint_no_te}
  e_\text{viewpoint, no TE} \definedas \frac{\Big|\substack{\text{\#frames with misclassified} \\ \text{viewpoints}\\-\text{\#frames with viewpoint TE}}\Big|}{\text{\#frames}}\times100\%.
  \end{equation}
\end{Definition}

For trajectory estimation, each estimated trajectory is plotted on a 2-D plane with \emph{unitless} axes, and the starting location mapped to the origin. Along a trajectory, the estimated poses are reconstructed using Rogez et al.'s 3-D, 13-jointed skeletal models \cite{rogez08spatio}. The proximity of the estimated trajectories to the actual trajectories was assessed.

\subsection{Experiments with publicly available datasets}\label{sec:exp:pub}

In this group of experiments, we used two publicly available human motion datasets: (i) CMU Motion of Body (MoBo) \cite{gross01cmu} and (ii) HumanEva2 \cite{sigal09humaneva}. Both CMU MoBo and HumanEva2 are recorded indoors by a ground-based camera with a static background. For these datasets, background subtraction is used for foreground/background segmentation, and tested only with HOG features. 

From the CMU MoBo dataset, the image sequence for subject 4071 was selected, which shows the subject walking on a treadmill at a constant speed. Fig.~\ref{fig_cmu_mobo}(a) shows the original images, the segmented silhouettes and the estimated poses. The reconstructed trajectory in Fig.~\ref{fig_cmu_mobo}(b) was a straight path, but the orientation was skewed by $45^\circ$ due to the $45^\circ$ error in the first estimated viewpoint. Table~\ref{table_classifier_acc} shows that the dynamic classifier had significantly lower values for $e_\text{pose}$ and $e_\text{viewpoint}$ than $C_{64}$.

From the HumanEva2 dataset, the image sequence for subject S2 combo C2 was selected, which shows the subject rounding a left turn. Figure~\ref{fig_humaneva2}(b) shows the 3-D reconstruction of the estimated poses and trajectory. The viewpoint of the third skeleton is incorrectly estimated as its adjacent viewpoint ($+45^\circ$ error). However, when moving from the third skeleton to the fourth one, the viewpoint was corrected. Table~\ref{table_classifier_acc} shows that the dynamic classifier had significantly lower values for $e_\text{pose}$ and $e_\text{viewpoint}$ than $C_{64}$.

\begin{table*}[h!]
  \centering
  \caption{Estimation errors of $C_{64}$ and the dynamic classifier using HOG features.}
  \label{table_classifier_acc}
  \begin{tabular}{lcccccc}
  \hline\noalign{\smallskip}
  \multirow{2}{*}{Experiment/dataset} & \multirow{2}{*}{\begin{tabular}[c]{@{}c@{}}\#frames\end{tabular}}
                              & \multicolumn{2}{c}{$e_\text{pose, with TE}$}
                              & \multicolumn{2}{c}{$e_\text{viewpoint, with TE}$} \\ \cline{3-6} 
              &       & $C_{64}$  & Dynamic classifier  & $C_{64}$  & Dynamic classifier \\ \noalign{\smallskip}\hline\noalign{\smallskip}
  CMU MoBo              & 35    & 62.9\%    & \textbf{0}\%               & 14.3\%             & \textbf{5.7}\% \\ 
  HumanEva2             & 130   & 73.8\%    & \textbf{59.2}\%            & 31.5\%             & \textbf{19.2}\% \\ 
  Scenario 1 ($h=2$m)   & 250   & 48.8\%    & \textbf{36.8}\%            & 6.4\%              & \textbf{4.8}\% \\ 
  Scenario 2 ($h=10$m)  & 784   & 30\%      & \textbf{23.5}\%            & \textbf{11.9}\%    & 13\% \\ 
  Scenario 3 ($h=10$m)  & 1652  & 27.5\%    & \textbf{23.5}\%            & \textbf{16.2}\%    & 16.9\% \\ \noalign{\smallskip}\hline
  \end{tabular}
\end{table*}

\begin{table*}[h!]
  \centering
  \caption{Estimation errors of the dynamic classifier using HOG/CNN features on UAV-captured videos.}
  \label{table_classifier_acc_hog_vs_cnn}
  \begin{tabular}{lcccccc}
  \hline\noalign{\smallskip}
  \multirow{2}{*}{Experiment} & \multirow{2}{*}{\begin{tabular}[c]{@{}c@{}}\#frames\end{tabular}}
                              & \multicolumn{2}{c}{$e_\text{viewpoint}$}
                              & \multicolumn{2}{c}{$e_\text{pose}$} \\ \cline{3-6} 
                  &       & CNN   & HOG     & CNN   & HOG \\ \noalign{\smallskip}\hline\noalign{\smallskip}
  Scenario 1 ($h=2$m), with TE    & 250   & 22.8\%        & \textbf{4.8}\%         & \textbf{34}\%  & 36.8\% \\ 
  Scenario 1 ($h=2$m), no TE      & 250   & \textbf{0}\%  & \textbf{0}\%           & 3.2\%          & \textbf{1.2}\% \\ 
  Scenario 2 ($h=10$m), with TE   & 787   & 44.5\%        & \textbf{13}\%          & 52.7\%         & \textbf{23.5}\% \\ 
  Scenario 2 ($h=10$m), no TE     & 787   & 15.7\%        & \textbf{0}\%           & 30.4\%         & \textbf{3.2}\% \\ 
  Scenario 3 ($h=10$m), with TE   & 1652  & 30.3\%        & \textbf{16.9}\%        & 41.5\%         & \textbf{23.4}\% \\ 
  Scenario 3 ($h=10$m), no TE     & 1652  & 16.9\%        & \textbf{0.4}\%         & 17\%           & \textbf{3.8}\% \\ \noalign{\smallskip}\hline
  \end{tabular}
\end{table*}

\subsection{Experiments with video captured from a UAV}\label{sec:exp:uav}

In this group of experiments, three video datasets representing three different scenarios were captured from a rotorcraft UAV --- specifically, a 3DR Solo --- in a slow and low-altitude flight mode. For recording videos, we use a GoPro Hero 4 black camera with an anti-fish eye replacement lens (5.4mm, 10MP, IR CUT) and a 3-axis Solo gimbal. The images were sampled at a rate of 30fps. In order to ease the segmentation process, the videos were recorded with an uncluttered background and with the human subject wearing dark clothes. The UAV-captured videos are segmented as described in Section~\ref{sec_segmentation}. These experiments were conducted using both HOG and CNN features.

\begin{figure}[ht]
\begin{center}
\includegraphics[width=\linewidth]{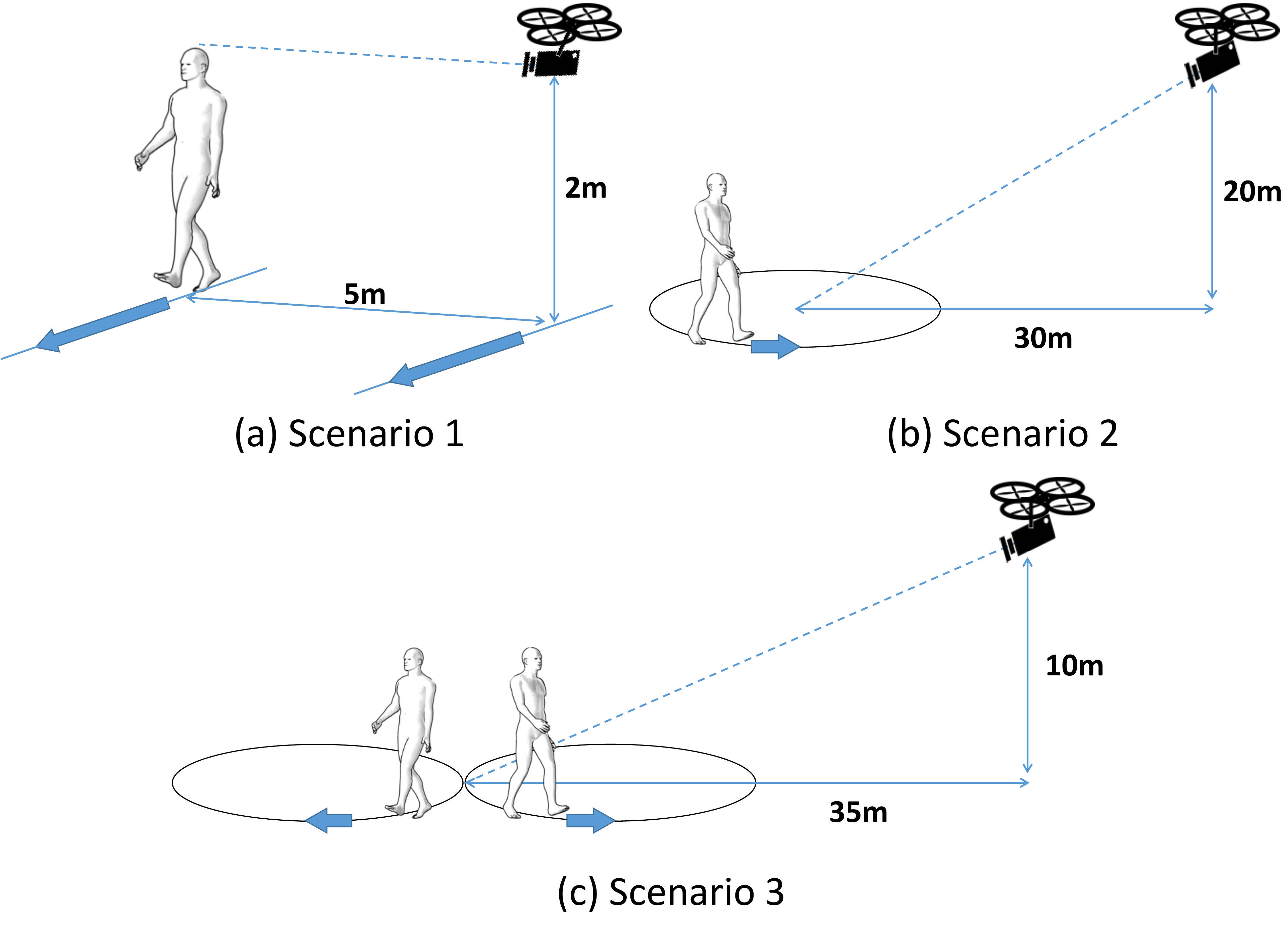}
\caption{How the videos for the three scenarios in Section~\ref{sec:exp:uav} were recorded by a UAV: (a) A human subject was filmed from his left walking along a straight line. The UAV moved in synchrony with the subject. (b) A human subject was filmed walking in a circle, while the UAV stays pointed at the center of the circle. (c) A human subject was filmed walking on a figure 8 shape, while the UAV stays pointed at the center of the shape.} 
\label{fig_scenarios}
\end{center}
\end{figure}

Certain assumptions were made to ease the coordinate transformation between the camera and the human subject:
\begin{itemize} 
\item    The human subject stands upright on flat ground. 
\item    The camera roll angle is zero.
\item    The roll, pitch and yaw angles of the UAV are zero during slow flight. Thus, the flight dynamics of the UAV has negligible effects on the camera elevation angle.
\item    The human subject is \emph{approximately} centered in the video.
\end{itemize}
These are valid assumptions in the case of an aerial platform designed to track human motion in a large field of view (see Fig.~\ref{fig_scenarios}). The camera elevation angle and height were directly recorded from the UAV control interface. The UAV was operated at a known ground distance (camera distance) from the human subject.

\begin{figure*}[ht]
\begin{center}
\includegraphics[width=0.8\linewidth]{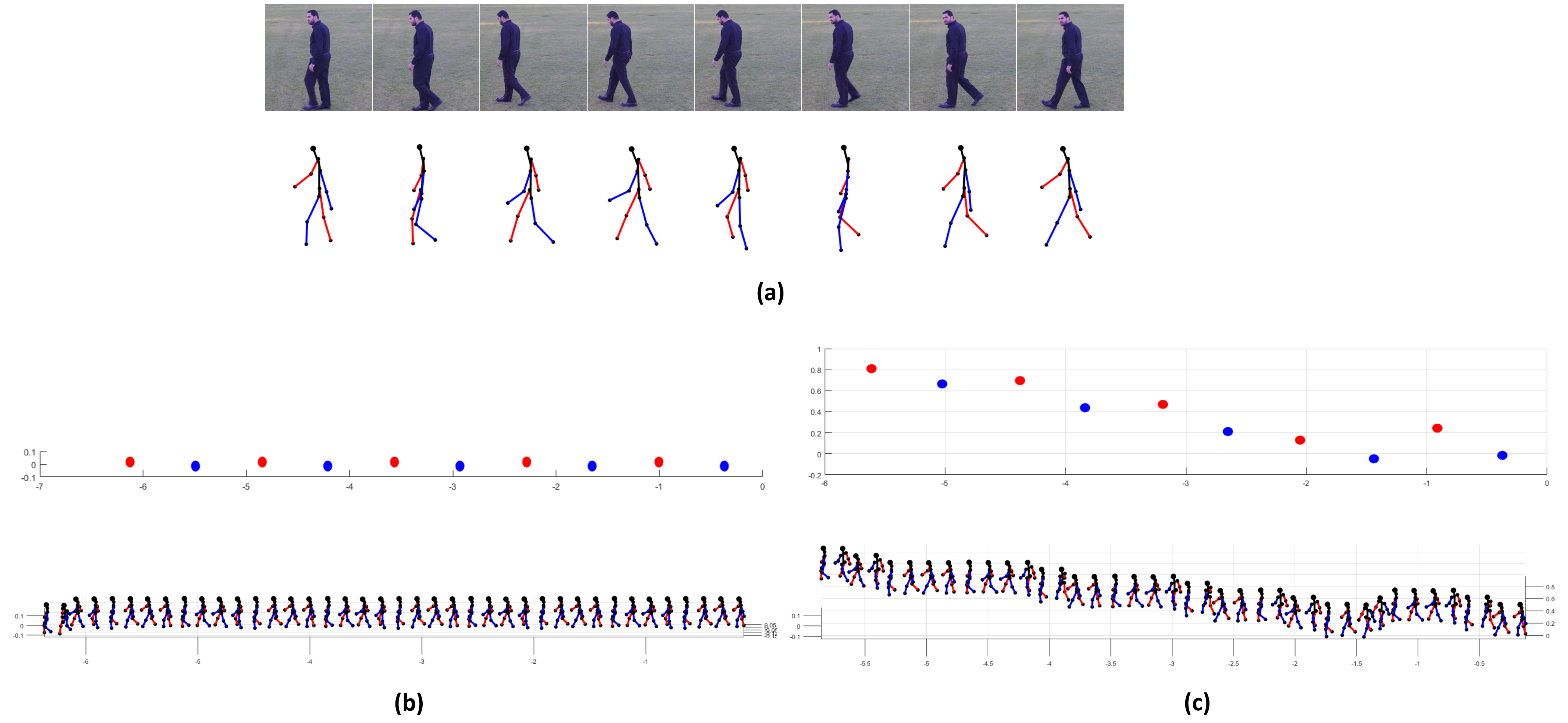}
\caption{Partial results for the Scenario 1 experiments, where a UAV follows a human subject on their left. (a) Some sample input images and their HOG-based estimated poses. (b) The estimated trajectory using HOG features. (c) The estimated trajectory using CNN features.} 
\label{fig_walk_straight}
\end{center}
\end{figure*}

\begin{figure*}[ht]
\begin{center}
\includegraphics[width=0.6\textwidth]{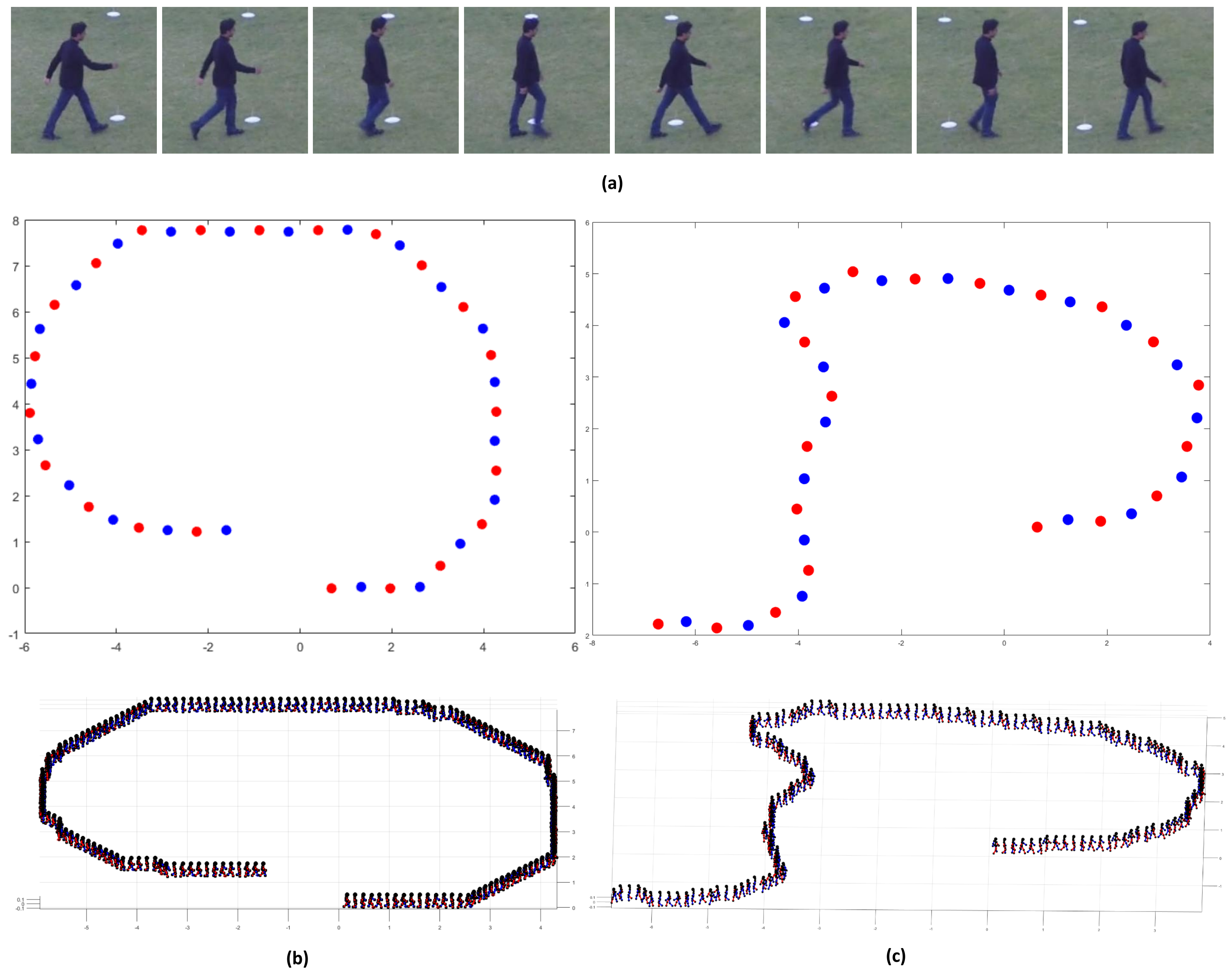}
\caption{Partial results for the Scenario 2 experiments: (a) The top row shows a series of cropped video frames. (b) The estimated trajectory using HOG features, where each dot marks where both feet touch the ground, with red representing right foot in front and blue representing left foot in front. 3-D reconstruction of the estimated poses and trajectory is also shown. (c) The estimated trajectory and 3-D reconstruction of the estimated poses using CNN features.} 
\label{fig_circle}
\end{center}
\end{figure*}  

\begin{figure*}[ht]
\begin{center}
\includegraphics[width=\textwidth]{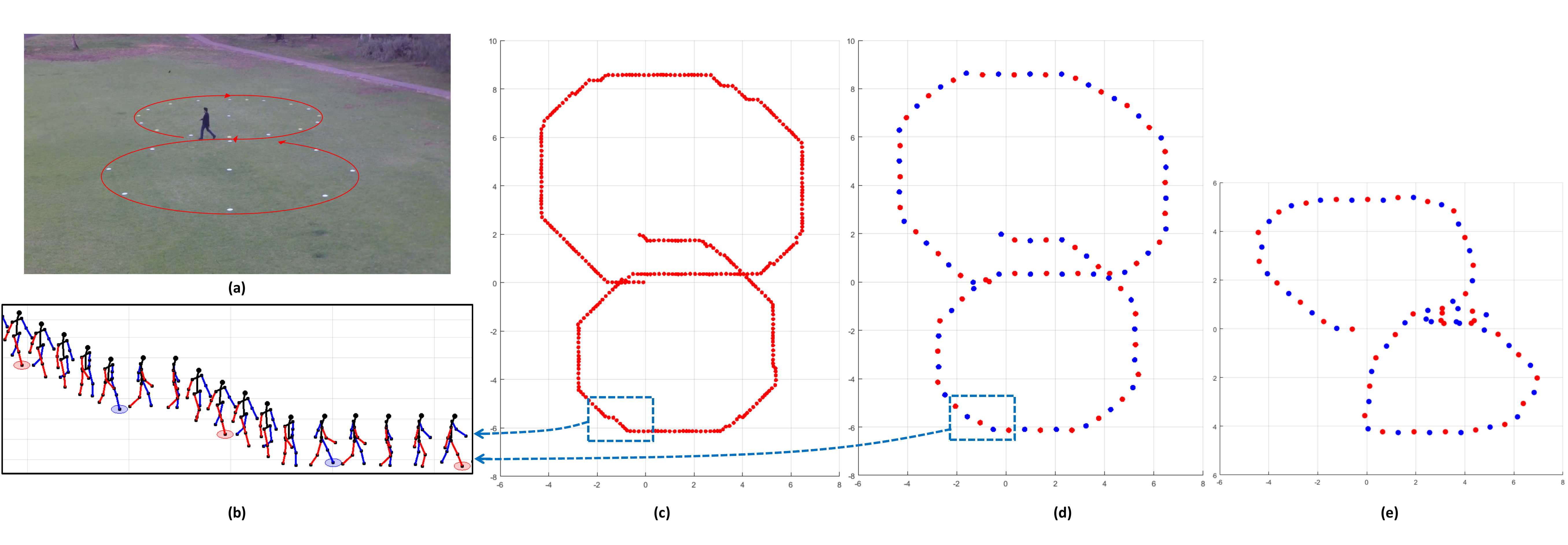}
\caption{Partial results for the Scenario 3 experiments ((b)-(d) correspond to HOG and (e) corresponds to CNN): (a) An original image of a human subject walking on a marked figure 8 path. (b) 3-D reconstruction of the estimated poses and trajectory, for a segment of the path shown in (c). (c) Tracking of the right foot of the human subject. Each point represents the estimated location of the right foot at each sub-step of the gait cycle. (d) Red and blue points represent where the right and left foot touches the ground respectively. (e) The estimated trajectory using CNN features.} 
\label{fig_shape8}
\end{center}
\end{figure*}

\begin{figure*}[ht]
\begin{center}
\includegraphics[width=\textwidth]{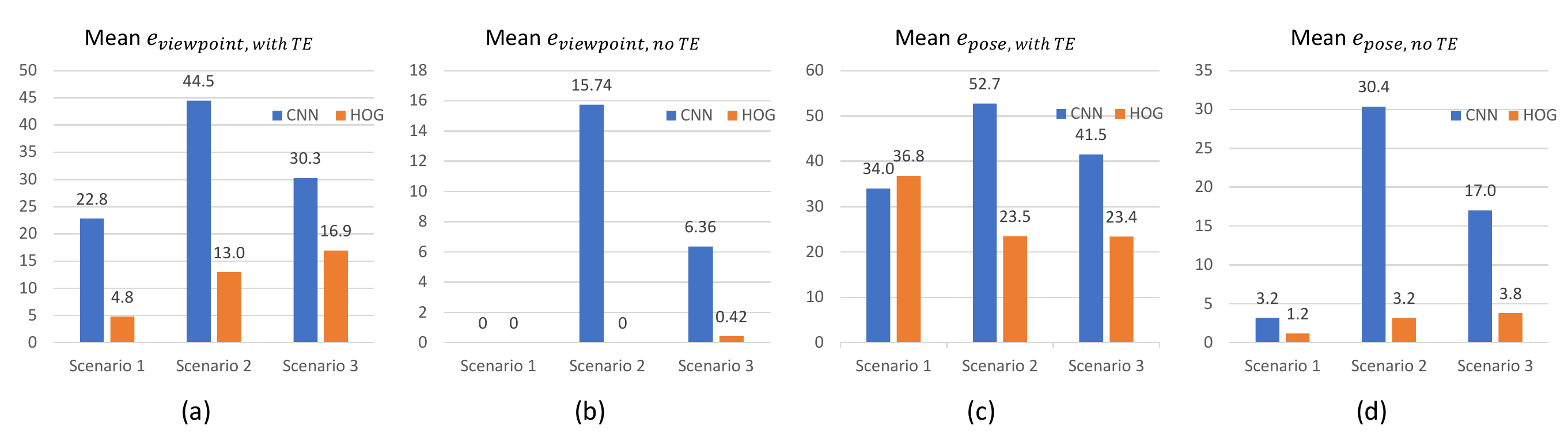}
\caption{Mean viewpoint and pose estimation error (\%).} 
\label{fig_tot_error_chart}
\end{center}
\end{figure*}
  
\begin{figure*}[t]
\begin{center}
\includegraphics[width=\textwidth]{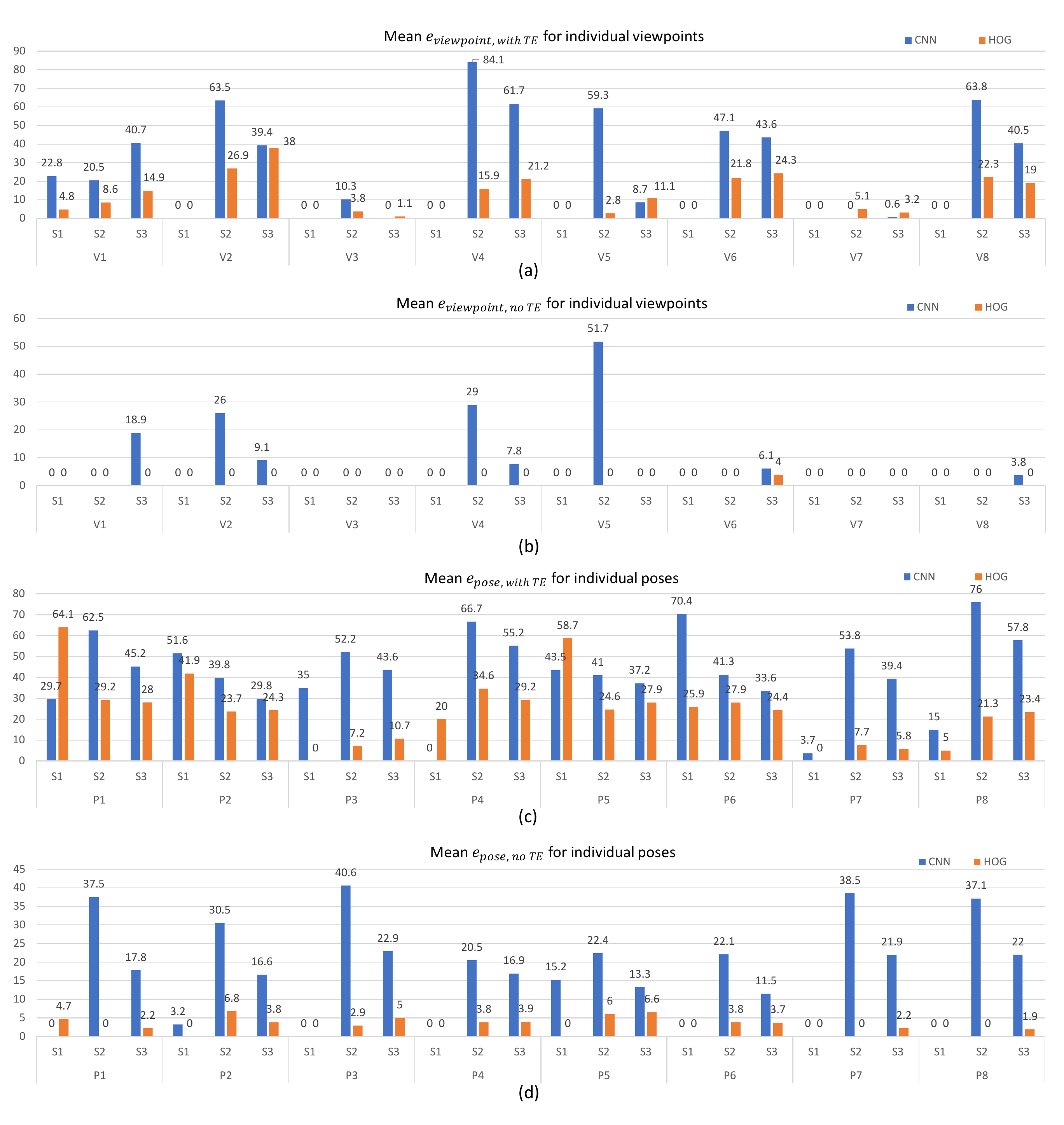}
\caption{Mean viewpoint and pose estimation error (\%) for each viewpoint and pose. S$_1$-S$_3$ denote Scenarios 1--3 respectively.} 
\label{fig_ind_error_chart}
\end{center}
\end{figure*}
  
\begin{figure*}[t]
\begin{center}
\includegraphics[width=\textwidth]{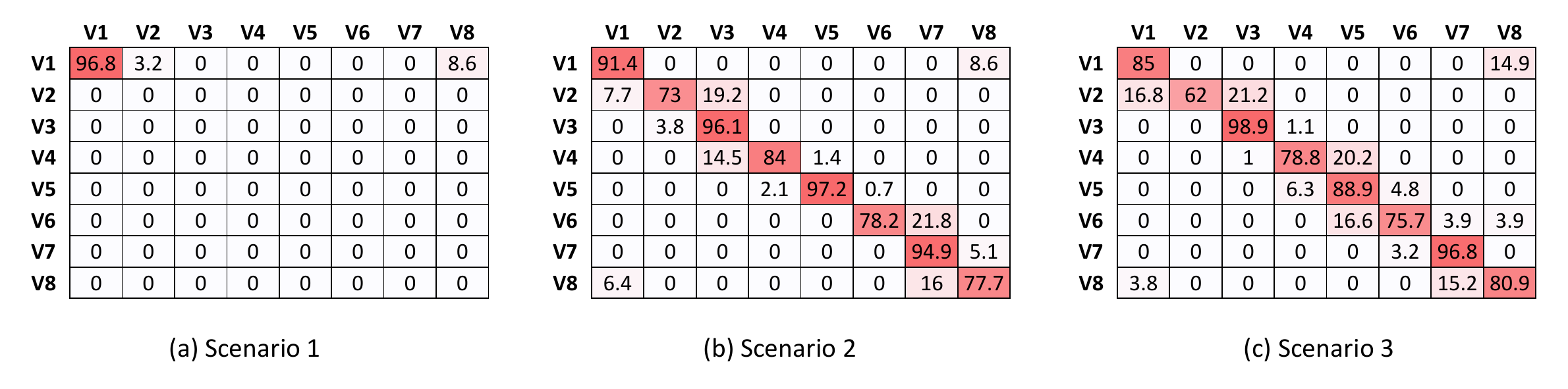}
\caption{Viewpoint confusion matrices for Scenarios 1--3 using HOG features.} 
\label{fig_conf_mat}
\end{center}
\end{figure*}

\subsubsection{Scenario 1}
As depicted in Fig.~\ref{fig_scenarios}(a), a human subject was filmed from his left walking along a straight line, by a UAV moving in synchrony. To achieve synchrony, the UAV was manually operated to maintain roughly the same speed as the subject, at a constant ground distance of 5m from the subject. The camera was horizontal and 2m above ground. Here are the findings:
\begin{itemize}
  \item Table~\ref{table_classifier_acc} shows that the dynamic classifier had lower values for $e_\text{pose}$ and $e_\text{viewpoint}$ than $C_{64}$.
  \item When comparing HOG and CNN in Table~\ref{table_classifier_acc_hog_vs_cnn} and Figs.~\ref{fig_tot_error_chart}--\ref{fig_ind_error_chart}, significantly lower values for $e_\text{viewpoint}$ and slightly higher values for $e_\text{pose}$ can be seen for HOG. Zero to very low errors are observable once the transitional errors were removed.  
  \item Figure~\ref{fig_walk_straight} shows a successful 3-D reconstruction of the estimated poses and trajectory for a segment of the path using HOG and separately CNN features.
  \item The confusion matrix in Fig.~\ref{fig_conf_mat}(a) shows viewpoint confusion was rare and confined to a neighbor of the true viewpoint.
\end{itemize}

\subsubsection{Scenario 2}
As depicted in Fig.~\ref{fig_scenarios}(b), a human subject was filmed walking on a marked circle by a UAV pointing at the center of the circle. The camera was 30m from the center of the circle and 10m above ground. Here are the findings:
\begin{itemize}
  \item Table~\ref{table_classifier_acc} shows that the dynamic classifier had a significantly lower $e_\text{pose}$ than $C_{64}$ and a slightly higher $e_\text{viewpoint}$ than $C_{64}$.  
  \item Table~\ref{table_classifier_acc_hog_vs_cnn} shows that CNN gives higher estimation errors than HOG does. The errors for CNN dropped significantly upon removal of the transitional errors.
  \item Figure~\ref{fig_circle}(b) shows the HOG-based estimated trajectory is approximately circular, as is the true trajectory. It also shows a 3-D reconstruction of the estimated poses and trajectory, with a small number of visibly wrong viewpoints. Figure~\ref{fig_circle}(c) shows the trajectory estimated using CNN features failed to follow the ground truth in the second half of the circular path.
  \item The confusion matrix in Fig.~\ref{fig_conf_mat}(b) shows viewpoint confusions are confined to neighbors of the true viewpoints. The lowest confusion rates are associated with the diagonal viewpoints $V_2$, $V_4$, $V_6$ and $V_8$; whereas, high classification accuracy has been recorded for $V_1$, $V_3$, $V_5$ and $V_7$. The reason is viewpoints $V_1$, $V_3$, $V_5$ and $V_7$ correspond to the front, back and side views. These four viewpoints suffer minimal self-occlusions in the silhouettes compared to the others, and hence, provide better image details. 
\end{itemize}

\subsubsection{Scenario 3}
As depicted in Fig.~\ref{fig_scenarios}(c), a human subject was filmed walking on a marked 8-shaped path by a UAV pointing at the center of the path, which was created by joining two circles of radius 5m. The walk starts and ends at the same point in the marked path. The camera distance from the middle of the path, where the two circles meet, was 35m and the camera height was 10m.
Perspective distortion of the video frames was negligible due to the small elevation angle, and so, the video frames were segmented without perspective correction. 
Here are the findings:
\begin{itemize}
  \item Table~\ref{table_classifier_acc} shows that the dynamic classifier has a significantly lower $e_\text{pose}$ than $C_{64}$ and a comparable $e_\text{viewpoint}$ to $C_{64}$. 
  \item Table~\ref{table_classifier_acc_hog_vs_cnn} shows CNN gave higher estimation errors than HOG does. These results are consistent with those for Scenarios 1--2.
  \item Figure~\ref{fig_shape8}(b) shows a 3-D reconstruction of the estimated poses and trajectory using HOG features for a segment of the path, which contains some visibly wrong viewpoints. However, Figs.~\ref{fig_shape8}(c) and \ref{fig_shape8}(d) show the estimated trajectory approximates the a figure 8 well. In Fig.~\ref{fig_shape8}(e), results for CNN approximately reflect the shape of the path, but both types of errors $e_\text{pose}$ and $e_\text{viewpoint}$ are significantly higher than those for HOG.
  \item The confusion matrix in Fig.~\ref{fig_conf_mat}(c) shows most of the viewpoint confusion was confined to neighbors of the true viewpoints. The worst confusion rate was associated with viewpoint $V_2$, whereas the highest classification accuracy was recorded for $V_3$. Generally, self-occlusions and loss of limb details are comparatively mild in $V_1$, $V_3$, $V_5$ and $V_7$, and hence they had the lowest confusion rates. Nevertheless, confusion rates depended largely on individual body dynamics.
\end{itemize}

\begin{table*}[t]
\centering
\caption{Estimation errors of $C_{64}$ and the dynamic classifier for perspective-distorted and perspective-corrected videos. Here, ``PD'' and ``PC'' refer to ``perspective distorted'' and ``perspective corrected'' respectively. }
\label{table_perspective_levels}
\begin{tabular}{llcccccc}
\hline\noalign{\smallskip}
\multicolumn{2}{c}{\multirow{2}{*}{Scenario 2}} & \multirow{2}{*}{\begin{tabular}[c]{@{}c@{}}\#frames\end{tabular}} 
                                                  & \multicolumn{2}{c}{$e_\text{pose, with TE}$}
                                                  & \multicolumn{2}{c}{$e_\text{viewpoint, with TE}$} \\ \cline{4-7} 
\multicolumn{2}{c}{}                      &                       & $C_{64}$  & Dynamic classifier  & $C_{64}$    & Dynamic classifier \\ \noalign{\smallskip}\hline\noalign{\smallskip}
                $h=10$m  & No distortion  & 787                   & \textbf{30}\%     & \textbf{23.5}\%   & \textbf{11.9}\%   & \textbf{13}\% \\ \noalign{\smallskip}
\multirow{2}{*}{$h=20$m} & PD             & \multirow{2}{*}{784}  & \textbf{37.2}\%   & \textbf{22.1}\%   & \textbf{18.9}\%   & \textbf{17.2}\% \\  
                         & PC             &                       & 49.9\%            & 39.9\%            & 20.9\%            & 20.4\% \\ \noalign{\smallskip}
\multirow{2}{*}{$h=30$m} & PD             & \multirow{2}{*}{810}  & 57.9\%            & 56.7\%            & 28.9\%            & 44.8\% \\  
                         & PC             &                       & \textbf{42.5}\%   & \textbf{40.6}\%   & \textbf{25.7}\%   & \textbf{37.2}\% \\ \noalign{\smallskip}
\multirow{2}{*}{$h=40$m} & PD             & \multirow{2}{*}{817}  & 68.4\%            & 74.4\%            & 38\%              & 42.6\% \\  
                         & PC             &                       & \textbf{53.5}\%   & \textbf{37.3}\%   & \textbf{30.5}\%   & \textbf{24.8}\% \\ \noalign{\smallskip}\hline
\end{tabular}
\end{table*}

\subsection{Experiments with perspective distortion}\label{sec:exp:pd}

\begin{figure*}[t!]
\begin{center}
\includegraphics[width=\textwidth]{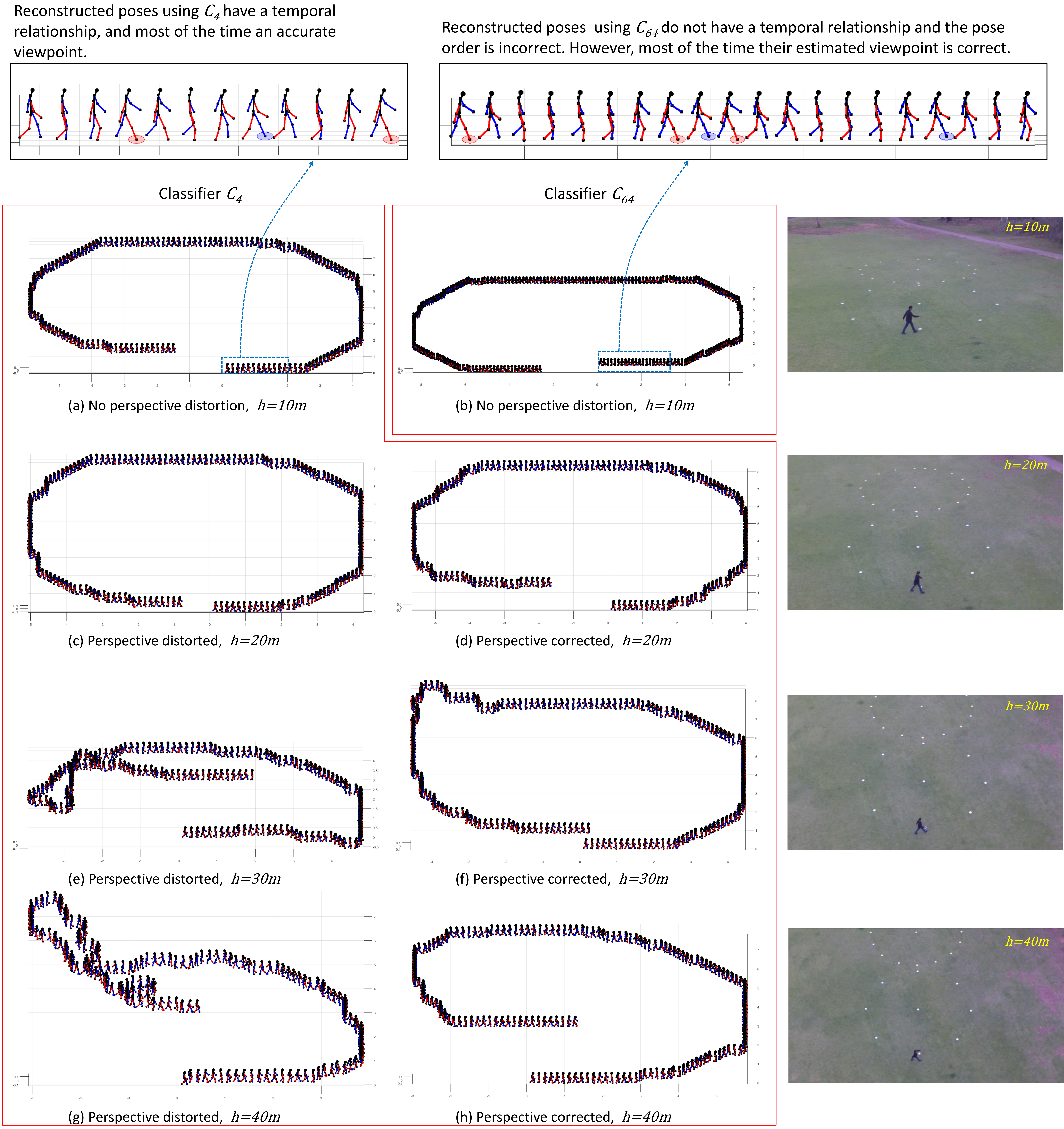}
\caption{Reconstructed poses and their trajectories correspond to four different heights are shown (using HOG features). The zoomed sections from graph (a) and (b) show the 3-D reconstruction of poses for the first $50$ frames using classifiers $C_4$ and $C_{64}$ respectively. Graph (b) does not give reliable pose or trajectory estimation due to the independently estimated poses of $C_{64}$. However, its viewpoint estimation is approximately equal to $C_4$. That is the reason it gives a similar trajectory to $C_4$. Graphs (c) to (h) present the effect of perspective distortion compensation on trajectory estimation for three different heights using the $C_4$ classifier. This trajectory estimation comparison can not be done with $C_{64}$ because it does not provide an ordinal relationship between poses. An original image captured at the respective height is included at the end of each row.} 
\label{fig_perspective_comparison}
\end{center}
\end{figure*}

This group of experiments was conducted using HOG features to analyze the effect of perspective distortion in detail. These experiments were extensions of the Scenario 2 experiments discussed in the previous subsection. In addition to 10m, the UAV was flown at heights of 20m, 30m and 40m (see Fig.~\ref{fig_perspective_comparison}). The lowest height of 10m caused negligible perspective distortion, but at $h=40$m ($\phi=53.1^\circ$), the video suffers from severe perspective distortion. The main observations are:
\begin{itemize}
  \item In terms of pose estimation accuracy, perspective correction helped the dynamic classifier, but not $C_{64}$, which was significantly worse than the dynamic classifier.
  \item In terms of viewpoint estimation accuracy, perspective correction helps the dynamic classifier much more than it helped $C_{64}$. 
  \item The advantage of perspective correction was more pronounced on more distorted videos.
  \item The advantage of perspective correction was more pronounced for the dynamic classifier than $C_{64}$.
\end{itemize}

Table~\ref{table_perspective_levels} once again confirms the advantage of the dynamic classifier over $C_{64}$, which does not take into account the ordinal relationship between poses. The advantage was more pronounced for more distorted videos, provided perspective correction was applied.

\section{Discussion}\label{sec_discussion}

Our discussion pertains to the dynamic classifier, HOG features, CNN features, perspective correction, limitations of the approach and considerations for practical implementation. 

\textbf{Dynamic classifier}. A drawback of the dynamic classifier is its dependence on accurate initial estimation. The solution given here is to use a multiclass classifier for the initialization, namely $C_{64}$, that recognizes all pose-viewpoint pairs. However, like all classifiers, $C_{64}$ sometimes makes mistakes, throwing the $C_4(\cdot,\cdot)$ classifiers off-course. A potential improvement is to re-initialize the dynamic classifier (see Algorithm.~\ref{alg_flowchart}) periodically.

\textbf{HOG features}. HOG features are traditionally considered to be handcrafted features, and in some domains, they have been replaced by CNN features. HOG cells in the literature do not capture additional information compared to CNN and they are significantly different features. HOG features are based on the weighted gradients in a HOG cell which represents the orientation of the edge lines. HOG are low-level features while CNN are high-level features with the ability to adapt to the task at hand during training.

However, given the robustness achieved in these multiclass classification experiments, HOG features outperform CNN features in pose and trajectory estimation. In our experiment, we extracted HOG features from a silhouette and CNN features from a color image. Our observation for the overall robustness of HOG features is that it is dependent on silhouettes and hence significantly on edges. However, segmentation of aerial images (for HOG) is very challenging due to the varying resolution and background, and can benefit from the latest advances in semantic segmentation.

\textbf{CNN features}. The accuracy of CNN-based feature extraction depends on many factors such as the neural network model, nature of the original training dataset and complexity of the test image. The followings are the possible reasons why we achieved a lower accuracy for CNN compared to HOG:
\begin{itemize}
  \item In the HOG approach, all the images are silhouettes and the features are formed from the edge details. In contrast, the CNN is sensitive to high-level features such as texture, background, face and gender, in addition to edges, and this can cause overfitting. 
  An overfitting model learns the noise and random details in training data in addition to the targeted details. A similar observation of HOG features outperforming CNN features in classification due to overfitting by the latter has been reported in \cite{sentas18performance}. Techniques such as dropout \cite{srivastava14dropout} and DeCov regularizer \cite{cogswell15reducing} have been proposed to reduce overfitting and increase generalization. However, we did not apply these techniques, and the scope of our finding is limited to standard transfer learning. 
  
  \item Transfer learning is a successful approach for many computer vision-related problems \cite{chaturvedi15deep}, but it has some constraints from the base network when copying the first $n$ layers of the base network to the first $n$ layers of the target network (left frozen feature layers). As a result, the feature layers do not change during the training of the new task. A possible alternative is to use deep transfer learning (DTL) which offers more flexibility when extracting high-level features \cite{yosinki14how}. DTL can perform layer-by-layer feature transference to solve a target problem in either a supervised or unsupervised setting \cite{kandaswamy17multi}.
  
  \item In a CNN, the features detected by earlier layers include low-level image details such as edges and colors. However, in the later layers the features progressively become more specific to the object categories of the original dataset.  
  
  \item We used the original weights of AlexNet because our new dataset is very small compared to the original pre-trained dataset. This standard practice of not changing weights for a small dataset helps to reduce overfitting \cite{yosinki14how}. 
  
  \item In many computer vision applications, CNN features outperform low-level features when the neural network has been trained with a sufficiently large, application-specific dataset \cite{jain14modeep}. 
  On the other hand, HOG does not need such a large dataset to achieve high accuracy.
\end{itemize}

\noindent Considering the factors above, it is not surprising CNN was outperformed by HOG in our experiments.

\begin{figure}[ht]
\begin{center}
\includegraphics[scale=0.4]{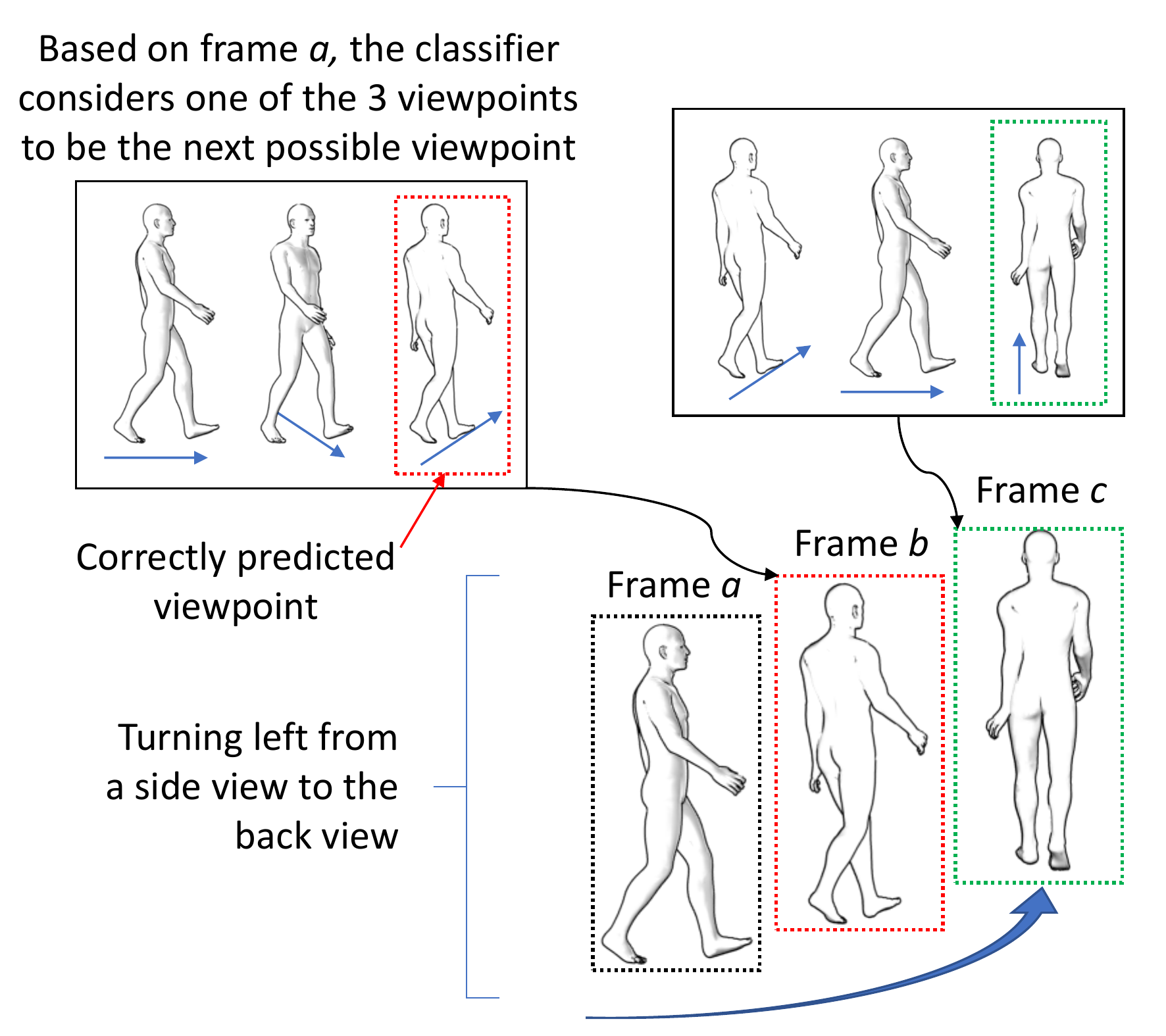}
\caption{An example of a subject turning left from a side view. 
In the classification stage, the first prediction is selected from the current and adjacent two viewpoints, each being $45^\circ$ different from the current viewpoint. The prediction comes from 4 classes (current pose and viewpoint, and the next pose with 3 possible viewpoints). For a left turn, the most likely first prediction is the one which is $45^\circ$ left to the current viewpoint. Again, in the next correct prediction, the prediction comes from 4 classes which includes the back view.} 
\label{fig_left_turn}
\end{center}
\end{figure}

\textbf{Left-right ambiguity}. To illustrate how our algorithm handles left-right ambiguity (i.e., confusion between front and back views), we present an example in Fig.~\ref{fig_left_turn}. Here, we consider a subject turning left from a side view to the back view. Once the subject turns $90^\circ$ to the left, his shape should be identifiable as the back view rather than the front view. As demonstrated in the figure, the back view cannot be confused with the front view, although they are similar in shapes, because the 4 classes are confined to three adjacent viewpoints related to the left turn.

\textbf{Perspective correction}. The results presented in Section~\ref{sec:exp:pd} confirm the intuition that perspective correction is imperative for severely perspective-distorted videos. Our solution has problems with purely frontal or rear views, because frontal and rear silhouettes do not provide sufficient details for differentiating pose. A potential solution is provided by the mobility of the aerial platform itself. The UAV can be programmed to seek a good elevation angle and azimuth angle, before it starts analyzing the human subject's action. This will require control algorithms and machine intelligence that go beyond the scope of this work.

\textbf{Limitations of the approach}. In this study, we tried to validate the suitability of dynamic classifiers for perspective distorted image sequences. We limited our work to gait estimation. However, the dynamic classifiers can be extended to estimate complex human poses. Another limitation is our system cannot handle complex gait sequences like sharp turns, twists and walking backwards. These are possible extensions to the current system, and can be addressed in future work. We used the standard transfer learning approach with Alexnet. However, an application-specific deep transfer learning framework can offer more flexibility to fine tune the neural network model. Finally, our training dataset is relatively small. The accuracy and the robustness of the classifiers can be further improved by adding more diverse images to the training data.

\textbf{Practical implementation}. The original motivation for this work was to make UAVs intelligent enough to recognize human activities, so the question about whether the proposed solution can run on an embedded platform is relevant. The most computationally intensive components of the proposed solution include homography, human detection, HOG feature extraction and SVM classification. The most computationally intensive is SVM, but even this can be implemented on resource-constrained devices \cite{anguita07hardware}. Further efficiency is ensured by the fact that a single 4-class classifier needs to run after initialization (recall Algorithm.~\ref{alg_flowchart}). In conclusion, all the algorithmic components are practical for an embedded platform. Note that 3-D reconstruction of the estimated poses and trajectory is meant for visualizations, not embedded applications.

\section{Conclusion and future work}
\label{sec_conclusion}

As a first step toward solving the problem of estimating human pose and trajectory in monocular videos from an aerial platform, the paper presents a solution that consists of perspective correction by homography, HOG/CNN feature extraction and dynamic classifier selection. The dynamic classifier is the defining feature of our solution, consisting of a 64-class classifier (namely $C_{64}$) and 64 4-class classifiers. The dynamic classifier works in conjunction with (i) a state transition model for the pose and viewpoint; and (ii) an SVM-based ECOC framework, which reduces multiclass classification to a set of efficiently solvable binary classification sub-problems (see Sects.~\ref{sec_training}--\ref{sec_classifier_design}). Trajectory estimation is for the estimation of the shape of the path traversed by the human subject, and is dependent on viewpoint estimation (see Section~\ref{sec_traj}).

Experiments have been conducted with the CMU MoBo and HumanEva2 datasets and our own UAV-captured datasets, using $e_\text{pose}$ and $e_\text{viewpoint}$ as defined in Equations.~\eqref{eq:e_pose_te}--\eqref{eq:e_viewpoint_no_te} as performance measures. The performance measures were calculated using two alternative feature sets (HOG and CNN), and the accuracies were compared before and after removing the transitional errors. Results show that 
\begin{itemize}
  \item The dynamic classifier outperforms $C_{64}$.
  \item Classification errors in the confusion matrix are evidently confined to neighbors of the true viewpoints. This property of the dynamic classifier enables fast recovery from incorrect estimations. 
  \item HOG features, compared to CNN features, facilitate more accurate estimation.
  \item The more perspective-distorted a video is, the more necessary perspective correction is for reducing the estimation errors of the dynamic classifier.
  \item The proposed solution works well with both indoor and outdoor videos, and both ground videos and perspectively distorted aerial videos. 
  \item The estimated trajectories approximate the actual trajectories well. 
\end{itemize}

The solution proposed in this article is limited to estimating walking gaits. Our immediate plan is to extend the current work to the recognition of gestures performed during either walking or standing. Replacing the current HOG descriptors with Yang et al.'s flexible mixtures-of-parts model~\cite{yang11articulated} should provide a promising start.

\begin{acknowledgements}
This project was partly supported by Project Tyche, the Trusted Autonomy Initiative of the Defence Science and Technology Group (grant number myIP6780).
\end{acknowledgements}

\noindent\textbf{Compliance with Ethical Standards}\\

\noindent\textbf{Conflict of Interest} The authors declare that they have no conflict of interest.\\

\noindent\textbf{Informed Consent} The data collection was conducted under the approval of University of South Australia's Human Research Ethics Committee (protocol no. 0000035185).

\bibliographystyle{vancouver}
\bibliography{IEEEabrv,ref}   


\end{document}